\documentclass[10pt,twocolumn,letterpaper]{article}

\usepackage[pagenumbers]{cvpr} %

\usepackage{graphicx, calc}
\usepackage{amsmath}
\usepackage{amssymb}
\usepackage{booktabs}
\usepackage[linesnumbered,ruled,vlined]{algorithm2e}
\usepackage{multirow}
\usepackage{subcaption}
\usepackage{makecell}
\usepackage[export]{adjustbox}
\usepackage[dvipsnames]{xcolor}
\usepackage{enumitem}
\usepackage{tikz}
\usepackage{colortbl}
\usepackage{color,soul}

\newcommand\dunderline[3][-1pt]{{%
		\sbox0{#3}%
		\ooalign{\copy0\cr\rule[\dimexpr#1-#2\relax]{\wd0}{#2}}}}%
\newcommand{\mathi}[2]{{\definecolor{rulecolor}{named}{#1}%
		\dunderline{2pt}{#2}}}

\definecolor{cvprblue}{rgb}{0.21,0.49,0.74}
\usepackage[pagebackref,breaklinks,colorlinks,citecolor=cvprblue]{hyperref}

\usepackage[capitalize]{cleveref}
\crefname{section}{Sec.}{Secs.}
\Crefname{section}{Section}{Sections}
\Crefname{table}{Table}{Tables}
\crefname{table}{Tab.}{Tabs.}

\definecolor{rred}{RGB}{216, 27, 96}
\definecolor{rblue}{RGB}{30, 136, 229}
\definecolor{rgreen}{RGB}{28, 151, 77}
\definecolor{ryellow}{RGB}{255, 193, 7}
\definecolor{rorange}{RGB}{255,167,0}
\definecolor{americanrose}{rgb}{1.0, 0.01, 0.24}

\newcommand{\todo}[1]{}

\newcommand{\RCc}[1]{}
\newcommand{\RC}[1]{#1}
\newcommand{\FPc}[1]{}

\newcommand{\ILc}[1]{}
\newcommand{\IL}[1]{#1}

\definecolor{c1}{HTML}{2196f3}
\definecolor{c2}{HTML}{4caf50}
\definecolor{c3}{HTML}{ff9800}
\definecolor{c4}{HTML}{f44336}
\definecolor{c5}{HTML}{673ab7}

\newcommand{\conesquare}[0]{\textcolor{c1}{$\blacksquare$}}%
\newcommand{\ctwosquare}[0]{\textcolor{c2}{$\blacksquare$}}%
\newcommand{\cthreesquare}[0]{\textcolor{c5}{$\blacksquare$}}%
\newcommand{\cfoursquare}[0]{\textcolor{c4}{$\blacksquare$}}%
\newcommand{\cfivesquare}[0]{\textcolor{c3}{$\blacksquare$}}%

\newcommand{\acg}{ACG\xspace}
\newcommand{\ph}{PH\xspace}
\newcommand{\cgb}{CGB\xspace}
\newcommand{\sd}{SD\xspace}

\newcommand{\OS}{OS\xspace}
\newcommand{\os}{OpenSurfaces\xspace}

\newcommand{\condenseparagraph}[1]{\noindent\textbf{#1}\quad}

\newcolumntype{H}{>{\setbox0=\hbox\bgroup}c<{\egroup}@{}}
\newcommand{\STAB}[1]{\begin{tabular}{@{}c@{}}#1\end{tabular}}
\newcommand{\perfu}[1]{#1} %
\newcommand{\oc}[1]{\cellcolor{gray!20}}

\providecommand{\ie}{}
\renewcommand{\ie}{\textit{i.e.}\xspace}
\providecommand{\cf}{}
\renewcommand{\cf}{\textit{cf.}\xspace}
\providecommand{\eg}{}
\renewcommand{\eg}{\textit{e.g.}\xspace}
\providecommand{\wrt}{}
\renewcommand{\wrt}{{w.r.t.}\xspace}

\newcommand{\region}{\ensuremath{\mathbf{\mathcal{R}}}}

\definecolor{bleudefrance}{rgb}{0.19, 0.55, 0.91}
\definecolor{darkspringgreen}{rgb}{0.09, 0.45, 0.27}
\newcommand{\method}{\ensuremath{\color{darkspringgreen}\mathsf{Material~Palette}}\xspace}

\newcommand{\image}{\ensuremath{\mathbf{\mathcal{I}}}\xspace}
\newcommand{\promptTrain}{\ensuremath{\texttt{Prompt}\texttt{Train}}\xspace}
\newcommand{\promptsMaterial}{\ensuremath{\texttt{Prompts}\texttt{Gen}}\xspace}
\newcommand{\concept}{\ensuremath{S^*}\xspace}
\newcommand{\klass}{\ensuremath{S^c}\xspace}

\newcommand{\newparagraph}[1]{\vspace{0.4em}\noindent\textbf{#1}}
\newcommand{\fsource}{\ensuremath{f^\mathcal{S}}\xspace}
\newcommand{\ftarget}{\ensuremath{f^\mathcal{T}}\xspace}
\newcommand{\pbrsdname}{TexSD}
\newcommand{\pbrsd}{\textbf{\pbrsdname}\xspace}
\newcommand{\svbrdf}{SVBRDF}
\newcommand{\lpips}{LPIPS}

\newcommand{\icon}[1]{
    \makebox[5px]{\raisebox{-1px}{\includegraphics[width=8px]{figures/icons/#1.png}}}
}

\definecolor{palred}{HTML}{e41a1c} %
\definecolor{palblue}{HTML}{377eb8} %
\definecolor{palgreen}{HTML}{4daf4a} %
\definecolor{palpurple}{HTML}{984ea3} %

\makeatletter
\let\old@rule\@rule
\def\@rule[#1]#2#3{\textcolor{rulecolor}{\old@rule[#1]{#2}{#3}}}
\makeatother

\definecolor{rulecolor}{named}{black}

\title{Material Palette: Extraction of Materials from a Single Image}

\author{
Ivan Lopes$^1$ \quad Fabio Pizzati$^2$ \quad Raoul de Charette$^1$ \\
$^1$Inria \qquad $^2$University of Oxford\\
\url{https://astra-vision.github.io/MaterialPalette}
}

\begin{document}

\twocolumn[{
\renewcommand\twocolumn[1][]{#1}
\maketitle
\thispagestyle{empty}
\begin{center}
  \vspace{-0.3cm}
  \newcommand{\teaserwidth}{\textwidth}
  \centerline{
    \includegraphics[width=\teaserwidth]{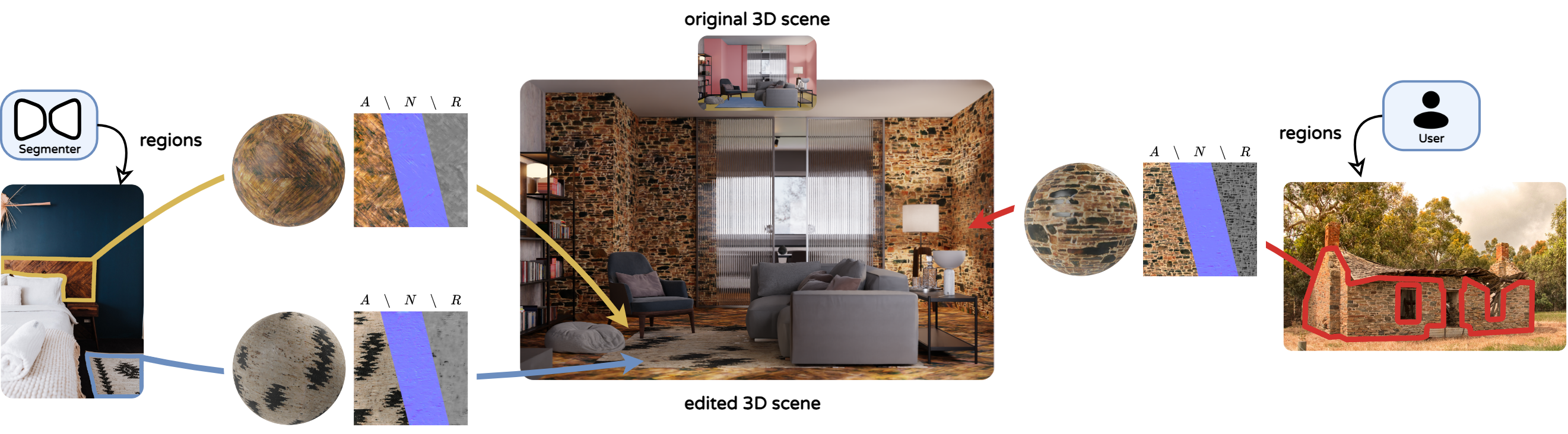}
    \vspace{-0.2cm}
    }
    \captionof{figure}{$\boldsymbol{\method}$. We introduce the task of material extraction from a real-world image \textit{without any prior knowledge}. 
    	Given an image as input (left and right), \IL{our method extracts Physically-Based Rendering (PBR) materials from input regions, which are either provided by a user (right) or output of a segmenter such as SAM~\cite{kirillov2023segment} (left). The extracted Spatially Varying BRDF (SVBRDFs) encode material intrinsics (\textbf{A}lbedo\textbackslash{}\textbf{N}ormal\textbackslash{}{}\textbf{R}oughness).} These can be reused for realistic material editing of 3D scenes (center).\todo{ivan x}%
     }%
    \vspace{-0.1cm}
  \label{fig:teaser}
 \end{center}
}]

\begin{abstract}
In this paper, we propose a method to extract Physically-Based-Rendering (PBR) materials from a single real-world image. We do so in two steps: first, we map regions of the image to material concepts using a diffusion model, which allows the sampling of texture images resembling each material in the scene. Second, we benefit from a separate network to decompose the generated textures into Spatially Varying BRDFs (SVBRDFs), providing us with materials ready to be used in rendering applications. Our approach builds on existing synthetic material libraries with SVBRDF ground truth, but also exploits a diffusion-generated RGB texture dataset to allow generalization to new samples using unsupervised domain adaptation (UDA). Our contributions are thoroughly evaluated on synthetic and real-world datasets. We further demonstrate the applicability of our method for editing 3D scenes with materials estimated from real photographs. The code and models will be made open-source.
\end{abstract}

\section{Introduction}
\label{sec:intro}

Whether it is a soft blanket, a rugged carpet, or a crumbling stone, humans can identify materials from a photograph. Besides geometry understanding, this ability derives from our sensing of how light interacts with materials, allowing us to identify the substance at stake without even touching it. 
\RC{In sciences, this has pushed research in spectrophotometry~\cite{germer2014spectrophotometry} or light sensing~\cite{bartels2019agile}, while in the arts, Vermeers and Caravaggio among others have used this long standing observation to convey the feeling of materials in their paintings.
Modern CG artists also deploy significant efforts to mimic realistic light-material interaction, through the design of Physically-Based Rendering materials (PBR).} 
While many libraries of material assets exist, no dataset can capture the true variety of real-world materials. What is more, capturing real-world materials is still a complex endeavor requiring special apparatus~\cite{asselin2020deep}.
In many scenarios, however, one may wish to estimate a material from a single RGB image, for example, to capture a unique marble stone in a visit or the fur of a wild animal from a souvenir photo.

Hence, we formulate the novel task of extracting PBR materials {from a single real-world image}, as shown in \cref{fig:teaser}. Given a set of regions, our method solves this task by generating corresponding textures along with their Spatially Varying BRDFs (SVBRDFs) \textit{without a priori knowledge} about the capturing viewpoint, scene geometry or lighting. This sets our work aside from the literature. 
We coined our method \method because, just like a painter would create their own color mix, one can create their palette of materials from their own photos (\cref{fig:teaser}, left and right). Moreover, extracted materials are readily usable for CG applications such as 3D renderings (\cref{fig:teaser}, middle).

There are major challenges in the estimation of PBR materials from just one RGB image, since single-view decomposition is highly ill-posed~\cite{deschaintre2018single}.
To address these hurdles, we rely on recent advances in text-to-image generation~\cite{rombach2022high,nichol2021glide,ramesh2022hierarchical} to disentangle the specific material appearance from the scene geometry and imaging conditions, allowing us to generate close-up tileable RGB textures of the materials in the scene. We further extract the PBR intrinsics of these diffusion-generated images, with a domain adaptation strategy that benefits from a novel synthetic dataset. \IL{Experiments show that \method outputs convincing results and performs better than baselines. The extracted materials closely resemble their real-world counterparts, which makes them usable for 3D scene editing.}

\noindent{}We contribute in the following ways:
\IL{\begin{itemize}[noitemsep,nolistsep]
	\item We formulate the novel challenging task of material extraction from a single real-world image. %
	\item We propose `\method', a method to extract materials within an image, operating in either a user-assisted or fully automated mode~(\cref{sec:automation}).
	\item Given an image region, we show how a finetuned text-to-image diffusion model can generate realistic tileable texture images~(\cref{sec:generative}) suitable for SVBRDF estimation~(\cref{sec:uda}).
	\item We provide a non-trivial evaluation pipeline to assess the quality of extracted PBR materials along with a novel prompt-generated dataset named~\pbrsd. 
	Experiments show our materials are close to those of real material datasets and readily usable for 3D editing. 
\end{itemize}
}

\section{Related works}
\label{sec:related}
To the best of our knowledge, we are the first to address end-to-end extraction of multiple materials from \IL{\textit{single} real-world} images but we cover literature connected to our task.

\paragraph{Single-image intrinsics decomposition.}
Long after the pioneering work of \cite{horn1974determining},
deep networks were leveraged for decomposition, exploiting their great pixel-wise estimation capabilities. %
Most early works focused on object-centric scenes with Lambertian assumption~\cite{tang2012deep}, user interaction~\cite{liao2019approximate}, or scene layers~\cite{innamorati2017decomposing}. To decompose in-the-wild objects, symmetry~\cite{wu2020unsupervised} or cross-instance~\cite{monnier2022share} constraints are applied, while~\cite{joy2022multi} requires the 3D mesh~\cite{joy2022multi}.
Holistic scene decomposition was addressed splitting albedo and shading~\cite{barron2013intrinsic,narihira2015direct,li2018cgintrinsics} also with the support of image-to-image translation~\cite{liu2020unsupervised}, or inverse rendering~\cite{sengupta2019neural}.
To account for spatially-varying lighting, some use mixture of illuminations or SVBRDF~\cite{barron2013intrinsic,zhou2019glosh,Garon_2019_CVPR,li2020inverse,li2022physically}. Notably, many of these works rely on estimated light sources and are designed for either indoor or outdoor scenes. 
Additionally, they capture the intrinsics of a scene image \IL{without distinguishing} between the materials present. We instead wish to extract the intrinsics of dominant materials.

\paragraph{Material and texture extraction.}
Typical material capture requires expensive multi-view~\cite{asselin2020deep} or polarized~\cite{deschaintre2021deep} apparatus. 
Many use synthetic data to train single-view SVBRDF estimation networks~\cite{deschaintre2018single}, often coupled with additional single-view data~\cite{gao2019deep,martin2022materia} or custom training strategies~\cite{li2017modeling,vecchio2021surfacenet,Deschaintre2020guided}. Importantly, all works mentioned \textit{require orthogonal close-up views} of the materials which is impractical for real scenes. UMat~\cite{rodriguez2023umat} uses a single image acquired with a flatbed scanner. \IL{Closest to our work is PhotoScene~\cite{yeh2022photoscene}, but it requires CAD inputs and is limited to a set of synthetic material graphs. Instead, we propose a single-image method targeting real-world material.
TexSynth~\cite{diamanti2015texsynth} provides a guided texture editing method but does not model explicitly the material.}\\
A connected field is texture extraction from real-world images. Note that while materials model light interaction, textures only describe the spatial arrangement of colors. A common strategy is to cluster the image textures and extend them to full resolution~\cite{rosenberger2009layered,li2022scraping} or apply dataset distillation~\cite{cazenavette2022wearable}. While we inspire from texture extraction methods, our task differs drastically as we seek to estimate the full SVBRDF -- not only the color.

\paragraph{Text-to-image generative models.}
Seminal works for text-to-image generations exploited conditional generative networks, allowing generation in constrained scenarios only~\cite{zhang2017stackgan,xu2018attngan,zhu2019dm}. Instead, training on billions of samples has been proven effective in generalizing on a wide range of prompts~\cite{ramesh2021zero,ramesh2022hierarchical}. To this extent, diffusion models are exploited for their stability at scale~\cite{ramesh2022hierarchical,nichol2021glide,saharia2022photorealistic,rombach2022high}, although adversarial-based methods are also used~\cite{sauer2023stylegan}. \IL{MatFuse~\cite{vecchio2023matfuse} and ControlMat~\cite{vecchio2023controlmat} recently adopted diffusion processes for material generation. We get inspiration from them while avoiding long training times.}

\begin{figure*}[t]
	\centering
	\includegraphics[width=\linewidth]{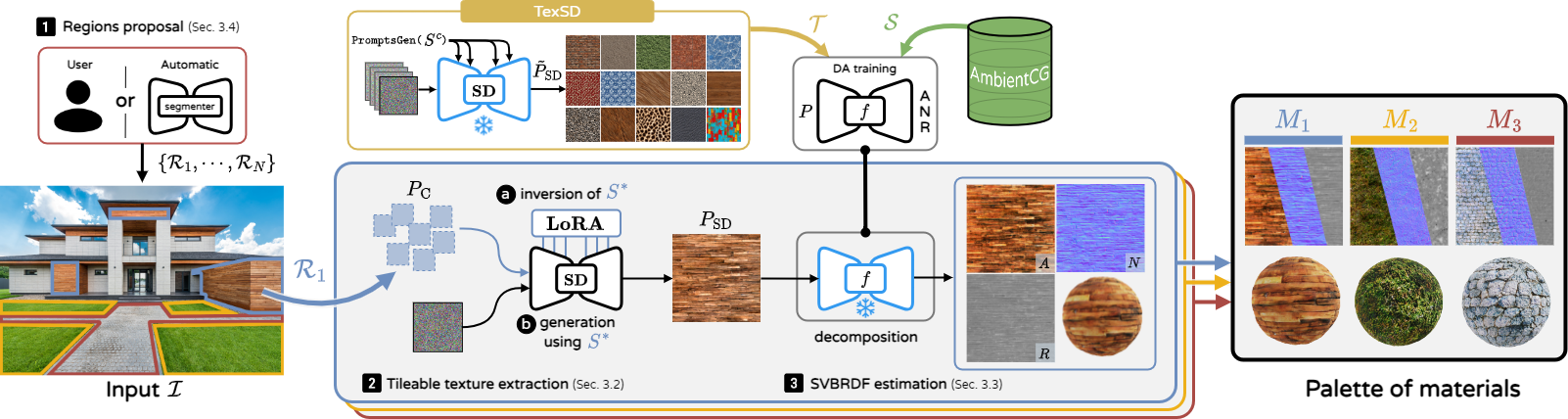}
	\caption{\textbf{$\boldsymbol{\method}$ pipeline}. From a single image \image (left) our method extracts the SVBRDF of dominant materials~(right). 
		Considering a set of regions $\{\region_1,\cdots,\region_N\}$ from a user or a segmenter~\icon{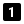}~(\cref{sec:automation}), we process each region $\region_i$ separately following two~steps. In \icon{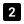}~(\cref{sec:generative}), we finetune Stable Diffusion~\cite{rombach2022high} on crops of the region $P_{\text{C}}$ to learn a concept $S^*$, which is later used to generate~a~texture image $P_\text{SD}$ resembling $P_{\text{C}}$. Then in~\icon{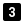}~(\cref{sec:uda}), these patches are decomposed into SVBRDF intrinsics maps~\mbox{(\textbf{A}lbedo, \textbf{N}ormal, \textbf{R}oughness)} using a multi-task network. Finally, \RC{the output is the palette} of extracted materials~$\{M_1,\cdots,M_N\}$ corresponding to input regions.%
    }\label{fig:method}
\end{figure*}

\section{Material Palette}
\label{sec:method}

\RC{Our method extracts Physics-Based Rendering (PBR) materials from regions of a real-world image.} Differently from approaches relying on close-up captures~\cite{deschaintre2018single,martin2022materia} or dedicated hardware~\cite{asselin2020deep}, the problem is much more challenging when given an in-the-wild image (Sec.~\ref{sec:statement}) with unknown scene lighting and geometry. Hence, we build on recent advances in vision-language models.

Given an input image \image and some input regions $\{\region_1, \cdots, \region_N\}$,~\method extracts a set of corresponding materials SVBRDF $\{M_1,\cdots, M_N\}$. 
Fig.~\ref{fig:method} illustrates the complete pipeline. 
For each region, we extract a texture approximating its material appearance using Stable Diffusion~\cite{rombach2022high}~(Sec.~\ref{sec:generative})\icon{two}.
Then, we rely on a domain-adaptive \RC{Spatially Varying BRDF (SVBRDF)} decomposition using our diffusion prompt-generated samples for \IL{generalizing} to the extracted textures~\icon{three}~(Sec.~\ref{sec:uda}). 
While our pipeline can rely on user input to define the image regions, we can also query any off-the-shelf segmenter~\icon{one}~(Sec.~\ref{sec:automation}).

\subsection{Problem statement}\label{sec:statement}

\begin{figure}
	\centering
	\footnotesize
	\setlength{\tabcolsep}{0.2pt}
	\newcolumntype{C}[1]{>{\centering\arraybackslash}m{#1}}
	\resizebox{0.98\columnwidth}{!}{
	     \begin{tabular}{l@{\hskip 0.4em}ccl@{\hskip 0.4em}ccc}
			& \multicolumn{3}{c}{Multi view training} & \multicolumn{3}{c}{Single view training} \\
			\cmidrule[1pt](r){2-4}\cmidrule[1pt](l){5-7}
	 	    \multicolumn{1}{c}{Image} & \textbf{A} & \textbf{N} & \multicolumn{1}{c}{\textbf{R}} & \textbf{A} & \textbf{N} & \textbf{R}\\
			\includegraphics[width=4.5em, height=4.5em]{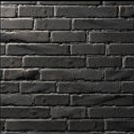} &
			\includegraphics[width=4.5em, height=4.5em]{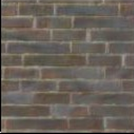} &
			\includegraphics[width=4.5em, height=4.5em]{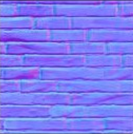} & 	
            \includegraphics[width=4.5em, height=4.5em]{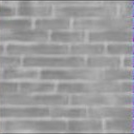} & 
			\includegraphics[width=4.5em, height=4.5em]{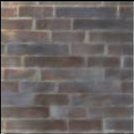} &
			\includegraphics[width=4.5em, height=4.5em]{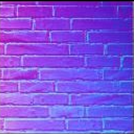} &
			\includegraphics[width=4.5em, height=4.5em]{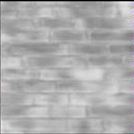} \\ 
			
		\end{tabular}	     
	}
	\caption{\textbf{Decomposition with known illumination.} Training on \acg~\cite{ambientcg} we note that even with known lighting, single-view leads to degenerated intrinsics  (\textbf{A},\textbf{N},\textbf{R}) due to viewpoint ambiguities.}\label{fig:decomptraining}
\end{figure}

Considering a typical intrinsics decomposition, an image $P$ with known illumination can be approximated as the result of a rendering operation $\tilde{\rho}(.)$ from SVBRDF maps $\tilde{M} = \{\tilde{A}, \tilde{N}, \tilde{R}\}$, being pixel-wise Albedo, Normals, and Roughness, respectively. This writes: $P = \tilde{\rho}(\tilde{M})$. 
Our goal is to learn the inverse rendering process with a neural network $f$ to predict $\tilde{M}$ from $P$. In details:
\begin{equation}
	f(P) = M = \{A, N, R\} \cong \tilde{M}  , 
\end{equation}
where $\tilde{M}$ is the ground truth SVBRDF.
A dataset with such labels is obtainable with expensive procedural generation of top-view materials or acquisitions in controlled scenarios~\cite{ambientcg}. We can train $f$, by rendering \textit{multiple views with known illuminations} $\tilde{\rho}_{1\dots{}n}$, and enforcing both a regression loss $\mathcal{L}_\text{reg}$ towards the ground truth maps and a multi-view rendering loss $\mathcal{L}_\text{ren}$ on the $n$ renderings~\cite{guarnera2016brdf}:
\begin{equation}
	\mathcal{L}_\text{reg} {=} ||M - \tilde{M}||_1,\,\,\,\,\,\, \mathcal{L}_{\text{ren}} {=} \sum_{i=1}^{n}||\tilde{\rho}_i(M) - \tilde{\rho}_i(\tilde{M})||_1\,.
	\label{eq:ren_rec_ren}
\end{equation}
After training, $f$ can be used to estimate $M$ from unseen images with unknown SVBRDF (\cref{fig:decomptraining}, multi-view). 
However, in our scenario we wish to estimate $M$ from specific regions of an \textit{in-the-wild} \RC{image with variable illumination and geometry, thus being shifted \wrt the training distribution of materials datasets.}
Besides, even \textit{assuming known illumination} a single view is ambiguous for surface normals and roughness estimation (\cref{fig:decomptraining}, single-view). 

Without any geometry or illumination priors, we tackle the problem per region by extracting tileable textures \IL{which are then decomposed into SVBRDF~(\cref{sec:generative}) while accounting for the domain gap~(\cref{sec:uda}).}

\subsection{Tileable texture extraction}
\label{sec:generative}

{Given an image \image and a material region $\region$, a naive approach to disentangle the appearance from the scene geometry/lighting would be to classify the material in $\region$ and use its label to generate patches with a \RC{text-to-image network~\cite{rombach2022high}}.} 
This would however fail to capture the fine-grained characteristics of the material in $\region$. 
{Considering for example the picture of a rundown house in~\cref{fig:teaser} (right), \RC{the label ``brick'' does not fully capture the intricate appearance of the dusty unaligned and weathered time-worn
stones}. Indeed, a mere classification of the material fails to encompass the complexity of the texture and its unique appearance.}

\RC{We formulate the problem as texture extraction from region \region, thus seeking to remove geometric distortion and lighting in \region{} by generating a flat texture image which we further decompose.} 
To do so, we build upon text-to-image models~\cite{rombach2022high}, exploiting their capabilities for disentangling semantics.
{Essentially, we finetune a text-to-image diffusion model~\cite{ruiz2023dreambooth} to encode the material depicted in $\mathcal{R}$ as a token. This allows us to generate a resembling tileable texture at any arbitrary resolution.}

{During finetuning (\mbox{\icon{two}\icon{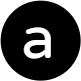}~in}~\cref{fig:method}), we extract crops $P_\text{C}$ from $\region$, utilizing them to map the material to a concept token~\concept. The aim is for~\concept to accurately describe the appearance of the specific material in \region, more faithfully than when using a class name.
In practice, we first finetune Stable Diffusion~\cite{rombach2022high} and learn \concept ~\cite{ruiz2023dreambooth,hu2022lora} using a \RC{single prompt template}, $\promptTrain{=}\textit{``an object with }{S^{*}}\textit{ texture''}$.}
Note, that the latter includes no information about the material class, removing \RC{needs for material labeling}. 

{During inference (\mbox{\icon{two}\icon{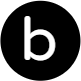} in}~\cref{fig:method}), we rely on different prompts \promptsMaterial chosen to enforce a texture-like appearance on a planar surface, such as ``\textit{realistic \concept texture in top view}''. 
\RC{The generative nature of the process let us generate not only one but a set of multiple textures from $\concept$, all resembling the material in \region. We rely on minimum \lpips~\cite{zhang2018unreasonable} w.r.t. crops of $\region$ to select the ad-hoc texture $P_{\text{SD}}$. We later detail the effect of prompts on the \IL{acquisition} of $\concept$ and generation of textures~(\cref{sec:exp-ablation}).}

Additionally, we follow ControlMat~\cite{vecchio2023controlmat} and adopt noise unrolling at inference to generate tileable textures. Different from ControlMat though, we are not conditioned on an input image at inference, but rather on~\concept which we can leverage to generate textures at \textit{any resolution}.

\subsection{SVBRDF estimation}%
\label{sec:uda}
\begin{figure}[t]
	\centering
	\includegraphics[width=\linewidth]{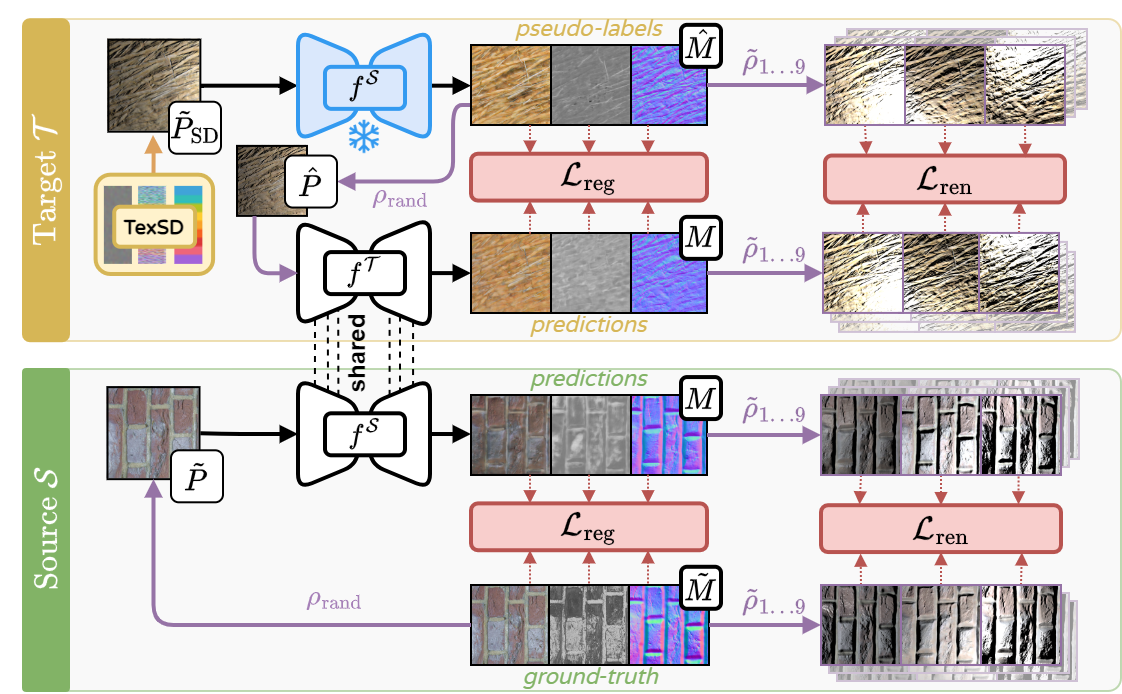}
    \caption{\textbf{SVBRDF Unsupervised Domain Adaptation.} We train a decomposition network $f$ on labeled SVBRDF materials $\mathcal{S}$ and unlabeled target data $\mathcal{T}$ \RC{from our novel \pbrsd dataset. Ultimately, $\mathcal{T}$ acts as a domain bridge between the SVBRDF dataset and the real domain,~\ie, patches generated from our extraction method~(\cf~\cref{sec:generative}).} We enforce both regression and rendering losses on $\mathcal{S}$ and $\mathcal{T}$ using pseudo-maps $\hat{M}$ extracted by the source-only model $f^{\mathcal{S}}$ (top). The final adapted model is denoted $f^{\mathcal{T}}$.
    }\label{fig:pm}
\end{figure}
We now seek to decompose each generated RGB-only texture from $P_{\text{SD}}$ into intrinsics \mbox{$M=\{A,R,N\}$}. 
From~Sec.~\ref{sec:statement}, a decomposition network $f$ can be trained on a SVBRDF dataset and used {on our generated textures~$P_{\text{SD}}$. 
Although these are much closer, than $\region$, to actual renderings of SVBRDF datasets\footnote{{By construction, $P_{\text{SD}}$ textures should be geometry- and lighting-free.}}, $f$ still suffers from a distribution shift.} We address this problem as an unsupervised domain adaptation (UDA) {$\mathcal{S}{\mapsto}\mathcal{T}$ where the source} domain $\mathcal{S}$ consists of materials with SVBRDF labels, and the target domain $\mathcal{T}$ is composed of diffusion-generated RGB textures.

\newparagraph{Source training.} We train our source model \fsource by enforcing a reconstruction loss on ground truth maps $\tilde{M}$ and a multi-view rendering loss with 9 lighting configurations\footnote{Light configurations are defined as angles $(\alpha, \phi)$ on the upper hemisphere, with $\alpha$ the light angle and $\phi$ the viewing angle. Following~\cite{deschaintre2018single}, we sample 6 symmetrical lighting/viewing angles to encourage specular and 3 uniformly sampled lighting/viewing angles to cover the parameter space.}, denoted {$\rho_{1\dots9}$}. Explicitly, we optimize $\fsource$ {by minimizing {$\mathcal{L}_{\text{reg}}$ and} $\mathcal{L}_{\text{ren}}$ defined in~\cref{eq:ren_rec_ren} with}:
\begin{equation}\label{eq:loss}
	\lambda\mathcal{L}_\text{reg}(M, \tilde{M}) + \mathcal{L}_{\text{ren}}(M, \tilde{M})\,
\end{equation}
\RC{where $M=\fsource(\tilde{P})$ and $\tilde{P}$ the rendering of $\tilde{M}$ with a random lighting $\rho_{\text{rand}}$. Ground truth maps $\tilde{M} \in \mathcal{S}$ are obtained from any SVBRDF library such as \acg~\cite{ambientcg}.}

\newparagraph{\pbrsd.} To bridge the gap with SVBRDF libraries, we first generate training material \RC{textures in the target domain} by prompting the text-to-image model with \promptsMaterial, \eg, ``\textit{realistic \klass texture in top view}'', \RC{replacing \klass with a class name. 
Notably, we \textit{do not} rely on finetuning, but {instead} only exploit the text-to-image capabilities of large diffusion models and their innate knowledge of material classes.} 
This allows us to construct a dataset, named~\pbrsd, of 9,000 textures generated from a set of 130 classes derived from \acg and ChatGPT proposals~\cite{chatgpt}. 
A schematic view is in~\cref{fig:method}~(block `\pbrsdname'), and details are in the supplementary.
{Crucially, despite the generation of multiple images per class, these texture images \RC{are not} multiple views of the same material instance. They are instead \textit{single-view} variations within a class.}

\newparagraph{Adaptation.} \RC{Equipped with \pbrsdname{} as target domain $\mathcal{T}$, we overcome the ill-posed single-view training (\cf~Fig.~\ref{fig:decomptraining}) drawing inspiration from pseudo-labels~\cite{lee2013pseudo}. We extract \textit{pseudo-decomposition maps} $\hat{M} = \{\hat{A}, \hat{N}, \hat{R}\}$ for all textures images \mbox{$\tilde{P}_\text{SD} \in \pbrsd$} by processing them with our source model $\fsource$. This enables pseudo multi-view training on $\mathcal{T}$ with \textit{only single view} images.} 

Hence, we adopt~\cref{eq:loss} to first train on $\mathcal{S}$ with ground truth $\tilde{M}$, and then finetune concurrently on $\mathcal{S}+\mathcal{T}$, using pseudo-labels $\hat{M}$ for $\mathcal{T}$. 
\RC{An illustration of the training process is shown in Fig.~\ref{fig:pm}. 
In both stages, SVBRDF inputs are rendered with random lighting conditions $\rho_\text{rand}$ to encourage invariance and robustness. 
At inference, we use $\ftarget$ to infer the SVBRDF of $P_{\text{SD}}$, leading to material maps $M_\text{SD}$.}

\subsection{Pipeline automation}
\label{sec:automation}
{Our pipeline is readily usable to extract materials from any region $\region$ of a real-world image. While 3D applications may benefit from user interaction to define $\region$, we also complement our pipeline with full automation.} 

To do so, we formalize the problem of defining regions $\{\region_1,\cdots,\region_N\}$ on real-world images as a 2D segmentation task. We integrate two segmentation models in our pipeline: the Segment Anything Model (SAM)~\cite{kirillov2023segment}, {a large-scale instance segmentation model,} and Materialistic~\cite{sharma2023materialistic}, a material selection method. In \cref{sec:exp-material}, we show how all of these region proposals lead to accurate material extractions.

\section{Experiments}
\label{sec:exp}

\RC{We study the performance of~\method along four main axes: \textbf{i)} Measuring the quality of our generated textures with respect to textures scraping techniques~(Sec.~\ref{sec:exp-texture}); \textbf{ii)} Quantifying our SVBRDF adaptation scheme on real material textures (Sec.~\ref{sec:exp-decomposition}); \textbf{iii)} Evaluating the quality of our extracted materials end-to-end (Sec.~\ref{sec:exp-material}); \textbf{iv)} Through exhaustive rendering of 3D scenes with our materials.
We also ablate our method in Sec.~\ref{sec:exp-ablation} and demonstrate the usage of our material palette for 3D editing in~\cref{sec:exp-app}.}

\subsection{Experimental details}
\label{sec:exp-details}
\condenseparagraph{Networks.}
We use Stable Diffusion~\cite{rombach2022high} for texture extraction, training a LoRA~\cite{hu2022lora} Dreambooth~\cite{ruiz2023dreambooth} for learning~\concept. Optimization times take around 3-5 min per learned~\concept on a Tesla V100-16GB. When learning~\concept, \promptTrain is set to ``\textit{an object with \concept texture}'' while for inference it is chosen randomly among~\promptsMaterial (see~\cref{fig:prompt-ablation}). To ensure tileability and high resolution generation, we roll the latent tensor by a random amount at every timestep of the diffusion process~\cite{vecchio2023controlmat}. We apply Poisson solving~\cite{perez2023poisson} to remove seams remaining on the borders. We directly sample textures up to 1024px \RC{while for higher resolutions, we batch-decode the latent code and blend overlapping patches using a weighted average}. For $f$, we use a multi-head CNN~\cite{lopes2023cross} \RC{with U-Net}~\cite{ronneberger2015u}, ResNet-101~\cite{he2015deep} backbone, and custom decoders \RC{with alternating upsample-conv layers}. Details are in the supplementary material.

\begin{figure}[t]
    \definecolor{mediumtealblue}{rgb}{0.0, 0.33, 0.71}
    \definecolor{mediumcandyapplered}{rgb}{0.89, 0.02, 0.17}
    \definecolor{mediumseagreen}{rgb}{0.24, 0.7, 0.44}
    \definecolor{mordantred19}{rgb}{0.68, 0.05, 0.0}
    
	\newcommand{\texrow}[1]{%
		\includegraphics[width=5em]{images/scrapingtex/#1_input.png} &%
		\includegraphics[width=5em]{images/scrapingtex/#1_patch.png} &%
		\includegraphics[width=5em]{images/scrapingtex/#1_WCT.png} &%
		\includegraphics[width=5em]{images/scrapingtex/#1_deeptexture.png} &%
		\includegraphics[width=5em,height=5em]{images/scrapingtex/#1_quilting.png} &%
		\includegraphics[width=5em]{images/scrapingtex/#1_psgan.png} &%
		\includegraphics[width=5em]{images/scrapingtex/#1_chosen.png} &%
		\includegraphics[width=5em]{images/scrapingtex/#1_xuetingli_zoom.png} &%

        \begin{tikzpicture} 
            \node[anchor=north west,inner sep=0] (image) at (0,0) {\includegraphics[width=5em]{images/scrapingtex/#1_2K_downsample.png}};
            \begin{scope}[x={(image.south east)},y={(image.north west)}]
                \draw[line width=1pt, mediumcandyapplered] (0.4,0.4) rectangle (0.4+0.5,0.4+0.5);
                \draw[line width=1pt, mediumtealblue] (0.4+0.125,0.4+0.125) rectangle (0.4+0.375,0.4+0.375);
            \end{scope}
        \end{tikzpicture} &%

        \adjincludegraphics[width=5em, trim={.4\width} {.1\height} {.1\width} {.4\height}, clip, cframe=mediumcandyapplered 1pt]{images/scrapingtex/#1_2K_downsample.png} &%

        \adjincludegraphics[width=5em, trim={.525\width} {.225\height} {.225\width} {.525\height}, clip, cframe=mediumtealblue 1pt]{images/scrapingtex/#1_2K_downsample.png}
        
	}
	\centering
	\footnotesize
	\newcolumntype{C}[1]{>{\centering\arraybackslash}m{#1}}
	\resizebox{1.0\linewidth}{!}{
		\tiny
		\setlength{\tabcolsep}{0.001\linewidth}
		\def\arraystretch{1.0}
		\begin{tabular}{ccHccccc ccc}
			& \multicolumn{5}{c}{Patch-based} & \multicolumn{5}{c}{{Region-based}} \\
			\cmidrule[1pt](r){2-6}\cmidrule[1pt](l){7-11}
			Image & Patch & WCT & DeepTex{\hspace{-0.1em}\cite{gatys2015texture}} & Quilting~\cite{efros2001image} & PSGAN~\cite{bergmann2017learning} & {Region} & Li \textit{et al.}~\cite{li2022scraping} &  \multicolumn{3}{c}{Ours ($\texttt{2048x2048}$)} \\
			\texrow{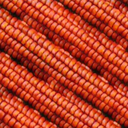}\\
			\texrow{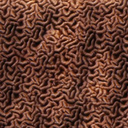}\\
			\texrow{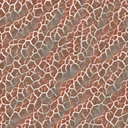}\\
			\texrow{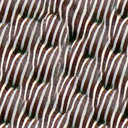}%
		\end{tabular}
	}
	\caption{\textbf{Textures extraction.} We compare texture extracted from natural images with 4 baselines being patch-based or region-based.  Different from baselines, our method is based on a learned concept~\concept which, when used for generating samples, corrects artifacts, is not limited to a fixed resolution and is fully tileable, resulting in homogeneous textures. For ease of comparison, we show outputs at $\texttt{2048x2048}$ along a {\color{mediumcandyapplered}\textbf{x2}} and {\color{mediumtealblue}\textbf{x4}} zoom. \textit{Images are downscaled for visualization purposes}.%
    }\label{fig:comparison-textures}
\end{figure}
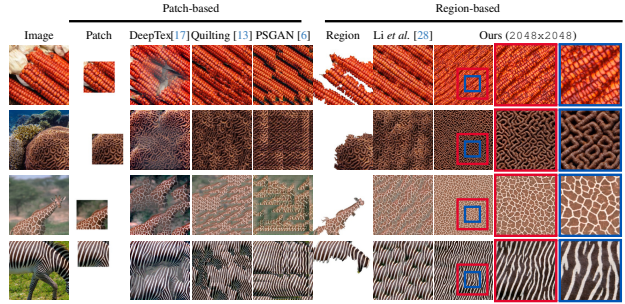

\condenseparagraph{Public datasets.}
We leverage three {SVBRDF} libraries (AmbientCG~\cite{ambientcg}, PolyHaven~\cite{polyhaven}, CGBookcase~\cite{cgbookcase}) and one material segmentation dataset (OpenSurfaces \cite{bell13opensurfaces}).\\
\textit{\textbf{AmbientCG}} (ACG) contains 2000 high-resolution PBR materials obtained from real-world captures with special apparatus, procedural generation, or image approximation. It includes around 50 material classes. We use the high-resolution 2k textures from ACG to train all source models.\\
\textit{\textbf{PolyHaven}} (PH) and \textit{\textbf{CGBookcase}} (CGB) are smaller libraries composed of 320 materials each. We use them as evaluating sets to validate our adaptation method.\\
\textit{\textbf{OpenSurfaces}} (OS) is an image dataset including dense material annotations. We use 14 overlapping classes with ACG and use a subset for end-to-end evaluation.\\

\condenseparagraph{\pbrsd dataset.} Our new \pbrsdname{} dataset (\cref{sec:uda}) for adaptation. It totals 9,000 textures generated at 1024x1024 by prompting Stable Diffusion~\cite{rombach2022high} with \promptsMaterial and 130 class names. We detail it in the supplementary.

\subsection{Texture extraction} 
\label{sec:exp-texture}
We showcase our text-to-image texture extraction~(\cref{sec:generative}) and existing techniques in~\cref{fig:comparison-textures}. 
We compare qualitatively with~\cite{li2022scraping} while also providing GAN-based baselines~\cite{bergmann2017learning}, methods inspired by style transfer~\cite{gatys2015texture} or image quilting~\cite{efros1999texture}. For a fair comparison, we show outputs from our method using the same regions as~\cite{li2022scraping}.\\
Even though~\cite{li2022scraping} outperforms older methods, it presents artifacts (last two rows) making images unsuitable for material extraction. Moreover, the extracted textures are non-tileable and entangle geometry and lighting. 
{In particular in the second row,~the texture of \cite{li2022scraping} replicates the shading of the input coral image.} 
Our method dramatically differs as it is able to map input images to plausible material textures, removing geometry and lighting while \textit{remaining devoid of artifacts}. 
Importantly, {we can generate any variations of \textit{tileable} samples, at \textit{any resolution}.} 
Ultimately, this shows the inadequacy of prior extraction methods for extracting tileable high-resolution texture patches suitable for decomposition and rendering purposes.

\begin{table}[t]
	\centering
	\small
	\setlength{\tabcolsep}{0.006\linewidth}
    \newcolumntype{P}[1]{>{\centering\arraybackslash}p{#1}}
	\newcommand{\vertical}[2]{\multirow{#1}{*}{\STAB{\rotatebox[origin=c]{90}{#2}}}}
    \newcommand{\darrow}[0]{\rotatebox[origin=c]{270}{$\Rsh$}}
	
    \resizebox{0.95\linewidth}{!}{
        \begin{tabular}{c!{\vrule width 1pt} r P{0.4cm}|cccH|cccH|c}

            \toprule
    		\multicolumn{1}{c}{} & &  & \multicolumn{4}{c|}{MSE ($10e1$) $\downarrow$} & \multicolumn{4}{c|}{SSIM $\uparrow$} & $\mathbf{\Delta}$\% $\uparrow$ \\[-0.1em]
    		\multicolumn{1}{c}{} & \scriptsize{method} & \scriptsize{DA} & \textbf{A} & \textbf{N} & \textbf{R} & I & \textbf{A} & \textbf{N} & \textbf{R} & I & \textbf{ANR} \\
    		\cmidrule{1-12}
    		
    		\vertical{4}{$\scriptsize{\text{\ph \textbf{ID}}}$}
    		& \scriptsize{Deep Materials~\cite{deschaintre2018single}} & & \perfu{0.264}& \perfu{0.380}& \perfu{0.453}& \perfu{0}& \perfu{0.379}& \perfu{0.235}& \perfu{0.358}& \perfu{0} &  \\  
    		& $\scriptsize{\mathcal{S}\text{ource-only}}$ & & \perfu{0.083}& \perfu{0.300}& \perfu{0.475}& \perfu{0.124}& \perfu{0.610}& \perfu{0.304}& \perfu{0.458}& \perfu{0.899}& \darrow \\
    		& \scriptsize{SurfaceNet~\cite{vecchio2021surfacenet}} & $\textbf{\checkmark}$ & \perfu{0.071}& \perfu{0.298}& \perfu{\textbf{0.427}}& \perfu{0.177}& \perfu{0.626}& \perfu{0.304}& \perfu{0.472}& \perfu{0.896}& +5.18 \\
    		& \oc{}\scriptsize{ours} & \oc{}$\checkmark$ & \oc{}\perfu{\textbf{0.069}}& \oc{}\perfu{\textbf{0.291}}& \oc{}\perfu{0.443}& \oc{}\perfu{\textbf{0.024}}& \oc{}\perfu{\textbf{0.630}}& \oc{}\perfu{\textbf{0.309}}& \oc{}\perfu{\textbf{0.476}}& \oc{}\perfu{\textbf{0.907}}&+\oc{}\perfu{\textbf{5.87}} \\
    		\cmidrule{2-12}
    		
    		\vertical{4}{$\scriptsize{\text{\cgb \textbf{ID}}}$}
    		& \scriptsize{Deep Materials~\cite{deschaintre2018single}} & & \perfu{0.590}& \perfu{0.465}& \perfu{1.940}& \perfu{0}& \perfu{0.392}& \perfu{0.228}& \perfu{0.346}& \perfu{0}  \\
    		& $\mathcal{S}\text{\scriptsize{ource-only}}$ && \perfu{0.098} & \perfu{0.221}& \perfu{\textbf{0.555}}& \perfu{x}& \perfu{0.662}& \perfu{0.437}& \perfu{0.476}& \perfu{x} & \darrow \\
    		& \scriptsize{SurfaceNet~\cite{vecchio2021surfacenet}} & $\checkmark$ & \perfu{0.101}& \perfu{0.230}& \perfu{0.615}& \perfu{0.077}& \perfu{0.657}& \perfu{0.445}& \perfu{0.471}& \perfu{0.900}& -3.00 \\
    		& \oc{}\scriptsize{ours} & \oc{}$\checkmark$ & \oc{}\perfu{\textbf{0.084}} & \oc{}\perfu{\textbf{0.219}}& \oc{}\perfu{0.588}&	\oc{}\perfu{x}&	\oc{}\perfu{\textbf{0.669}}& \oc{}\perfu{\textbf{0.457}}	&\oc{}\perfu{\textbf{0.482}} & \oc{}\perfu{\textbf{0.913}} & \oc{}\textbf{+2.62} \\
    		
            \cmidrule[0.5pt](r){1-12}
      
    		\vertical{4}{$\scriptsize{\text{\ph \textbf{OOD}}}$}
    		& \scriptsize{Deep Materials~\cite{deschaintre2018single}} & & \perfu{0.264}& \perfu{0.380}& \perfu{0.453}& \perfu{0}& \perfu{0.379}& \perfu{0.235}& \perfu{0.358}& \perfu{0} & \\
    		& $\mathcal{S}\text{\scriptsize{ource-only}}$ && \perfu{0.065}& \perfu{0.247}& \perfu{0.439}& \perfu{0.012}& \perfu{0.608}& \perfu{0.284}& \perfu{0.518}& \perfu{0.904}& \darrow \\
    		& \scriptsize{SurfaceNet~\cite{vecchio2021surfacenet}} & $\checkmark$ & \perfu{\textbf{0.053}}& \perfu{0.250}& \perfu{0.415}& \perfu{0.011}& \perfu{0.608}& \perfu{0.283}& \perfu{0.524}& \perfu{0.900}& +3.94 \\
    		& \oc{}\scriptsize{ours} & \oc{}$\checkmark$ & \oc{}\perfu{\textbf{0.053}}& \oc{}\perfu{\textbf{0.246}}& \oc{}\perfu{\textbf{0.409}}& \oc{}\perfu{\textbf{0.010}}& \oc{}\perfu{\textbf{0.618}}& \oc{}\perfu{\textbf{0.288}}& \oc{}\perfu{\textbf{0.531}}& \oc{}\perfu{\textbf{0.914}} & \oc{}\perfu{\textbf{+5.24}} \\
            \bottomrule

    	\end{tabular}%
     }\vspace{-0.5em}%
	\caption{\textbf{DA evaluation.} Performances on In-Distribution (ID, top) and Out-of-Distribution (OOD, bottom). $\Delta$ measures the relative performance over \{\textbf{A},\textbf{N},\textbf{R}\} {w.r.t.} the \acg-only model. DA refers to methods using domain adaptation strategies. Our adaptation succeeds at preserving generalization abilities on OOD samples, while SurfaceNet~\cite{vecchio2021surfacenet} has slightly lower gains on OOD.}
	\label{tab:quantitative_uda}
\end{table}

\begin{figure}[t]
	\newcommand{\dacomp}[1]{%
		\includegraphics[width=4.5em, height=4.5em]{images/acg_acgsd_comp/#1.png}%
		& \includegraphics[width=4.5em, height=4.5em]{images/acg_acgsd_comp/#1_acg_albedo.png} %
		& \includegraphics[width=4.5em, height=4.5em]{images/acg_acgsd_comp/#1_acg_normal.png} %
		& \includegraphics[width=4.5em, height=4.5em]{images/acg_acgsd_comp/#1_acg_roughness.png} %
		& \includegraphics[width=4.5em, height=4.5em]{images/acg_acgsd_comp/#1_acg_sphere.png} %
		& \includegraphics[width=4.5em, height=4.5em]{images/acg_acgsd_comp/#1_acgsd_albedo.png} %
		& \includegraphics[width=4.5em, height=4.5em]{images/acg_acgsd_comp/#1_acgsd_normal.png}%
		& \includegraphics[width=4.5em, height=4.5em]{images/acg_acgsd_comp/#1_acgsd_roughness.png}%
		& \includegraphics[width=4.5em, height=4.5em]{images/acg_acgsd_comp/#1_acgsd_sphere.png}%
	}
	\centering
	\footnotesize
	\setlength{\tabcolsep}{0.2pt}
	\newcolumntype{C}[1]{>{\centering\arraybackslash}m{#1}}
	\resizebox{0.98\columnwidth}{!}{
		\begin{tabular}{l@{\hskip 0.4em}cccc@{\hskip 0.4em}cccc}
			& \multicolumn{4}{c}{Source-only (ACG)} & \multicolumn{4}{c}{Ours $\text{\acg}\mapsto\text{\sd}$} \\
			\cmidrule[1pt](r){2-5}\cmidrule[1pt](l){6-9}
			\multicolumn{1}{c}{Image} & \textbf{A} & \textbf{N} & \textbf{R} & \multicolumn{1}{c}{3D} & \textbf{A} & \textbf{N} & \textbf{R} & \multicolumn{1}{c}{3D}\\
			\dacomp{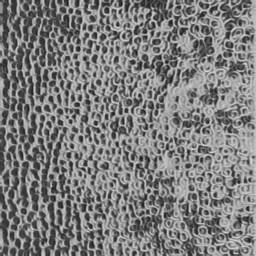}\\ %
			\dacomp{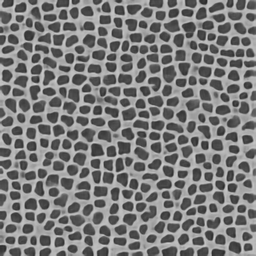}\\ %
			\dacomp{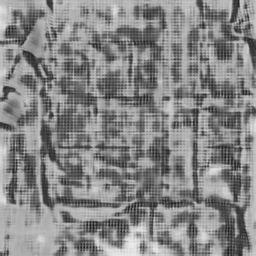}\\ %
		\end{tabular}
	}\vspace{-0.5em}
	\caption{\textbf{Qualitative results.} Comparison between source-only training (ACG) and our ($\text{\acg}\mapsto\text{\sd}$) adaptation on unseen SD samples. We notice that source-only overestimates N and R and exhibits color shift in A. This results in a lower-quality 3D render.}%
    \label{fig:qualitative-acg-sd}
\end{figure}

\subsection{SVBRDF decomposition}
\label{sec:exp-decomposition}
We now focus on our proposed UDA pipeline for decomposition (\cref{sec:uda}). Since our~\pbrsd dataset does not come with associated SVBRDF ground truths, we rather evaluate on two additional $\mathcal{S}{\mapsto}\mathcal{T}$ scenarios: $\text{\acg}{\mapsto}\text{\ph}$ and $\text{\acg}{\mapsto}\text{\cgb}$. This allows us to measure the adaptation effectiveness on the target set $\mathcal{T}$ {\wrt ground truth annotations}. We report standard metrics: Mean Squared Error ($MSE$~$\downarrow$) and Structural Similarity Index ($SSIM$~$\uparrow$)~\cite{wang2004image}. For a refined comparison, we evaluate common classes in $\mathcal{S}$ and $\mathcal{T}$ as In-Distribution (ID), resulting in 53 and 58 materials for \ph and \cgb, respectively.

We propose three baselines. First, we evaluate Deep Materials~\cite{deschaintre2018single} as an off-the-shelf decomposition network. We also compare with an \acg $\mathcal{S}\text{ource-only}$ model, serving as \textit{lower bound}. Then, we implement SurfaceNet~\cite{vecchio2021surfacenet} with our architecture and finetune the original \acg model for both SurfaceNet and ours. Results in~\cref{tab:quantitative_uda} (top) suggest that we improve consistently the decomposition. Considering the richer \ph material ontology, we also evaluate the 47 classes of materials Out-Of-Distribution (OOD). In Tab.~\ref{tab:quantitative_uda} we obtain only a low-performance OOD drop (bottom) compared to ID (top), {exhibiting better generalization than baselines.}
Furthermore, we show in~\cref{fig:qualitative-acg-sd} a visual comparison of `Ours $\text{\acg}{\mapsto}\text{\sd}$' vs `source-only (ACG)', on \pbrsd \textit{unseen} samples. It shows our adaptation better decomposes images, ultimately producing more realistic 3D renderings.

\begin{table}[t]
	\newcommand{\cred}[1]{\cellcolor{red!5}}
	\newcommand{\cgreen}[1]{\cellcolor{green!5}}
	\centering
	
	\resizebox{0.95\linewidth}{!}{%
		\begin{tabular}{cc|ccc}
			\toprule
			&&\multicolumn{3}{c}{LPIPS $\downarrow$}\\
			&\region & \textbf{A} & \textbf{N} & \textbf{R}\\
			\midrule
			\multicolumn{2}{c|}{\cred{}\textit{\footnotesize{upper bound}}} & \cred{}0.8288 & \cred{}0.5915 & \cred{}0.7255\\
			\midrule
			\multirow{3}{*}{\rotatebox[origin=c]{90}{Ours}} & \OS masks \cite{bell13opensurfaces} & \textbf{0.7959} & 0.5730 & 0.7142\\
			&SAM \cite{kirillov2023segment} & 0.8048 & 0.5692 & \textbf{0.7096}\\
			&Materialistic \cite{sharma2023materialistic} & 0.8077 & \textbf{0.5678} & 0.7169\\     
			\midrule
			\multicolumn{2}{c|}{\cgreen{}\textit{\footnotesize{lower bound}}} & \cgreen{}0.6789 & \cgreen{}0.4629 & \cgreen{}0.6843\\
			\bottomrule
		\end{tabular}\hspace{2em}%
		\begin{tabular}{cc|cc}
			\toprule
			&&\multicolumn{2}{c}{CLIP Classif. $\uparrow$}\\
			&$\rho(M)$& \textbf{top-1} & \textbf{top-5}\\
			\midrule
			& \acg & 47.78 & 85.88\\
			\midrule
			\multirow{3}{*}{\rotatebox[origin=c]{90}{Ours}}& \OS masks \cite{bell13opensurfaces} & 47.03 & 85.12\\
			& SAM \cite{kirillov2023segment} & 43.71 & 80.83\\
			& Materialistic \cite{sharma2023materialistic} & \textbf{50.89} & \textbf{86.82}\\		
			\bottomrule
		\end{tabular}%
	}
	\caption{\textbf{Resemblance to SVBRDF dataset.} We evaluate our extracted materials with various regions proposals w.r.t. materials from ACG (left). We also report zero-shot classification on ACG (right) which measures ability of CLIP to classify the class of the re-rendered material~\cite{radford2021learning}. 
		Both results demonstrate that our materials have coherent class-wise characteristics \textit{without class annotations}, irrespective of the segmenter used.}
\label{tab:quant-material-extraction}
\end{table}

\begin{figure*}[t]
	\centering
	\fontsize{75pt}{75pt}\selectfont
	\newcommand{\midrulesize}{10pt}
	
	\setlength{\tabcolsep}{.2em}
	\renewcommand{\arraystretch}{0}
	\newcommand{\imheight}{350px}
	\newcommand{\ballheight}{200px}
	\newcommand{\pic}[3]{\begin{tikzpicture}%
			\node[anchor=south west,inner sep=0] (image) at (0.0,0) {\includegraphics[height=\imheight#3]{#2}};%
			\begin{scope}[x={(image.south east)},y={(image.north west)}]
				\node[anchor=west] at (0.0,0.9) {\hspace{-0.35em}\textcolor{white}{\textbf{(#1)}}}; %
			\end{scope}
	\end{tikzpicture}}
	\resizebox{.99\linewidth}{!}{
		\begin{tabular}{cc @{\hskip .8em} cc @{\hskip .8em} ccc}
			\multicolumn{2}{c}{Materialistic} & \multicolumn{2}{c}{SAM} & \multicolumn{3}{c}{User-defined}\\[0.2em]
			\cmidrule[\midrulesize](lr){1-2}\cmidrule[\midrulesize](lr){3-4}\cmidrule[\midrulesize](lr){5-7}\\[0.2em]%
			\pic{a}{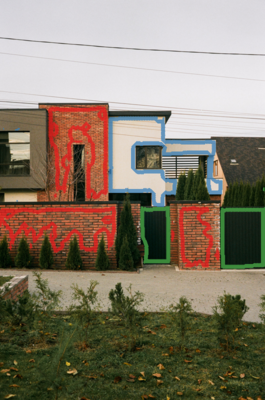}{}&%
			\pic{b}{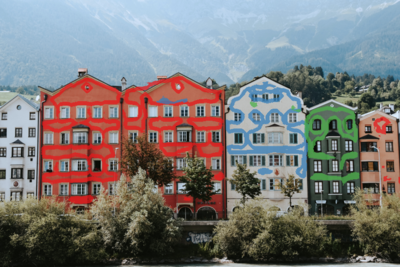}{}&%
			\pic{c}{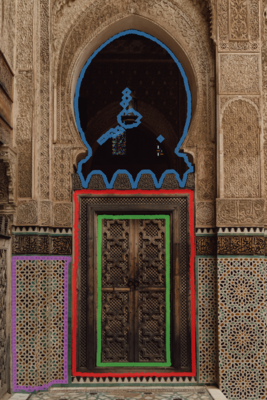}{}&%
			\pic{d}{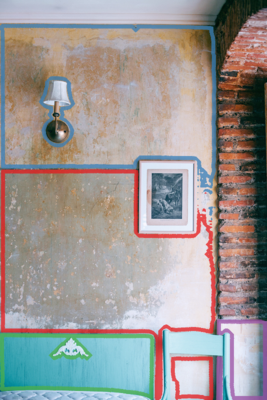}{}&%
			\pic{e}{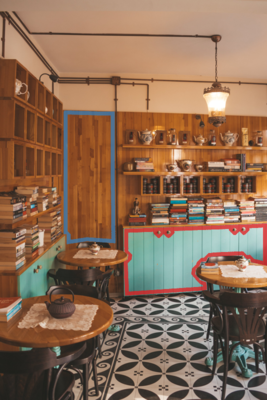}{}&%
			\pic{f}{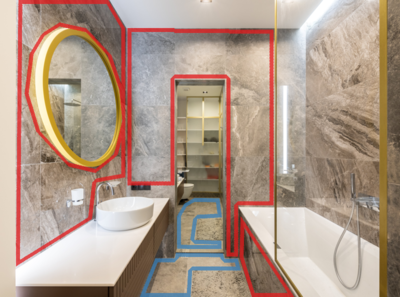}{}&%
			\pic{g}{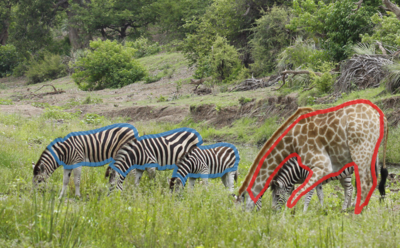}{}\\[0.2em]
			
			\includegraphics[height=\ballheight]{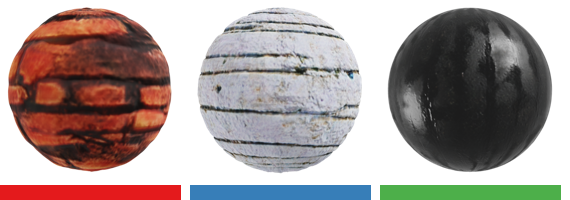}&%
			\includegraphics[height=\ballheight]{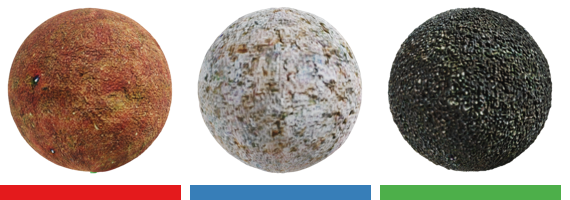}&%
			\includegraphics[height=\ballheight]{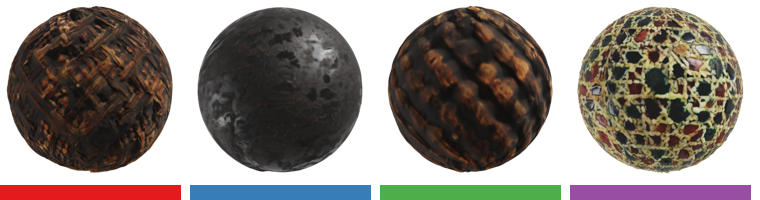}&%
			\includegraphics[height=\ballheight, trim=0 0 381px 0, clip]{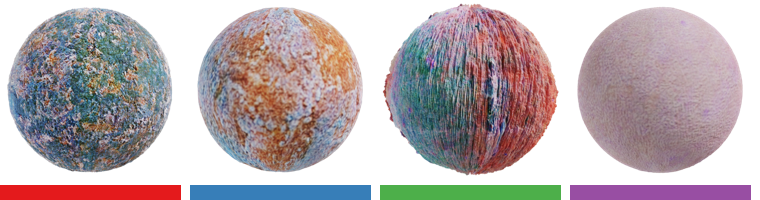}&%
			\includegraphics[height=\ballheight]{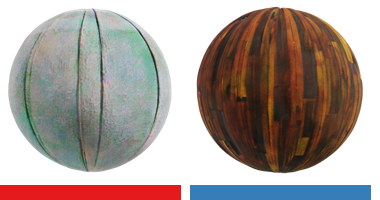}&%
			\includegraphics[height=\ballheight]{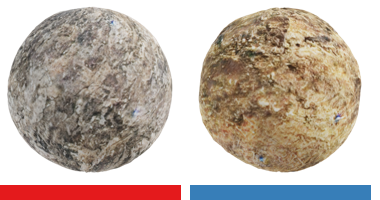}&%
			\includegraphics[height=\ballheight]{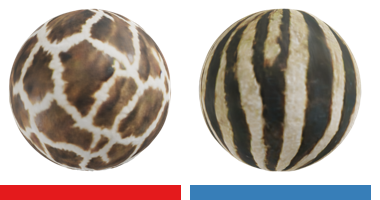}\\[0.6em]
			
			\pic{h}{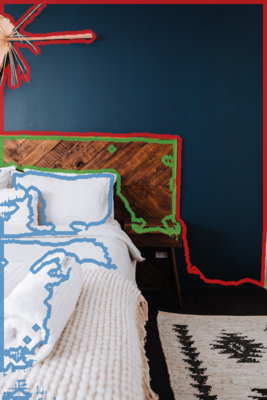}{}&%
			\pic{i}{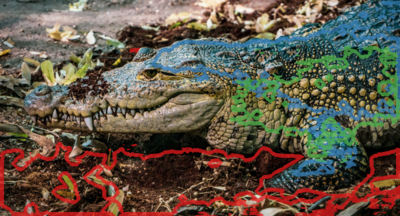}{}&%
			\pic{j}{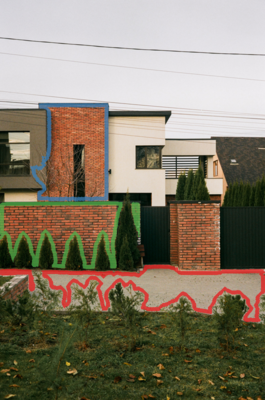}{}&%
			\pic{k}{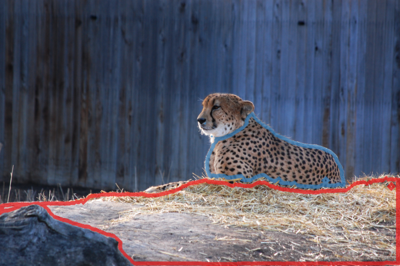}{, trim=0px 0 20px 0, clip}&%
			\pic{m}{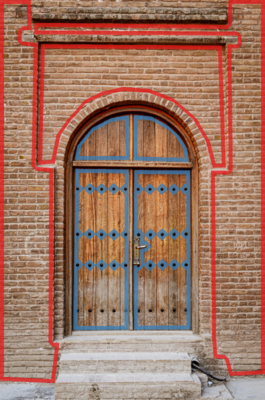}{}&%
			\pic{n}{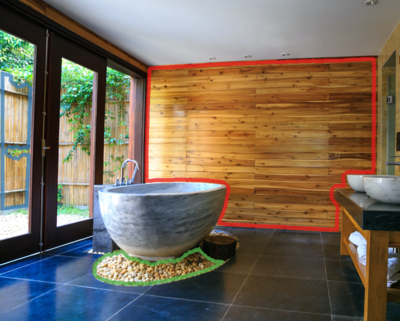}{, trim=0 0 0 0, clip}&%
			\pic{o}{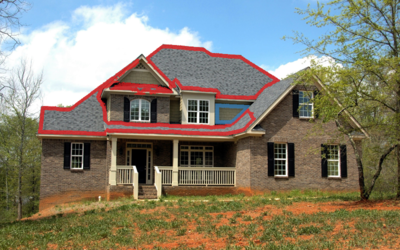}{, trim=0 0 20px 0, clip}\\[0.2em]
			
			\includegraphics[height=\ballheight]{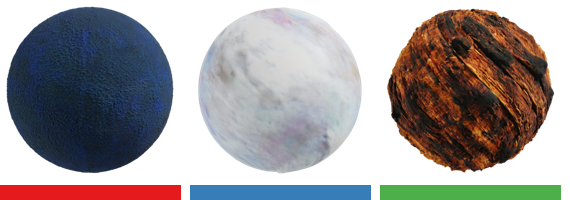}&%
			\includegraphics[height=\ballheight]{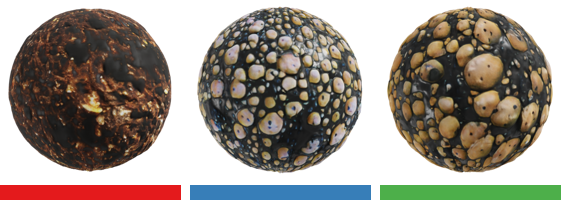}&%
			\includegraphics[height=\ballheight]{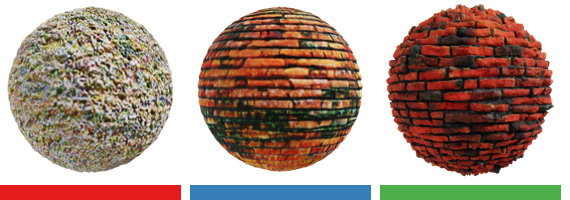}&%
			\includegraphics[height=\ballheight]{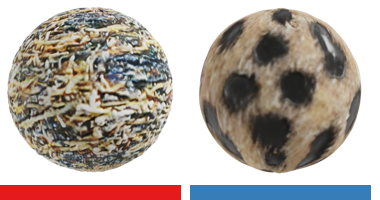}&%
			\includegraphics[height=\ballheight]{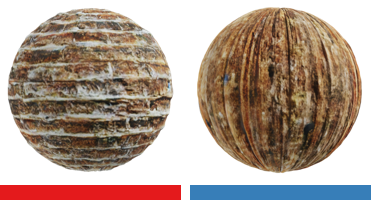}&%
			\includegraphics[height=\ballheight]{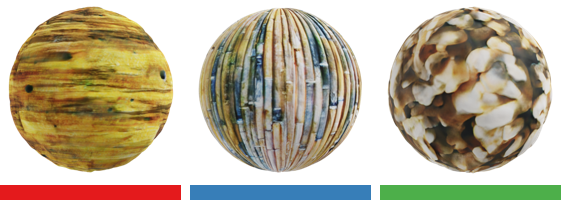}&%
			\includegraphics[height=\ballheight]{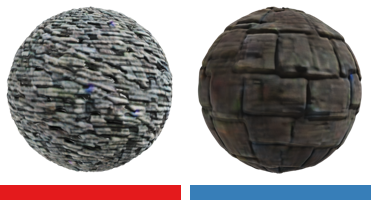}
		\end{tabular}%
	}%
	
	\caption{\textbf{Material palettes.} We show qualitative results of~\method on images gathered on the internet with regions extracted using Materialistic, SAM, or defined by the user. \RCc{Space to detail more here}
	}%
	\label{fig:palettes}
\end{figure*}
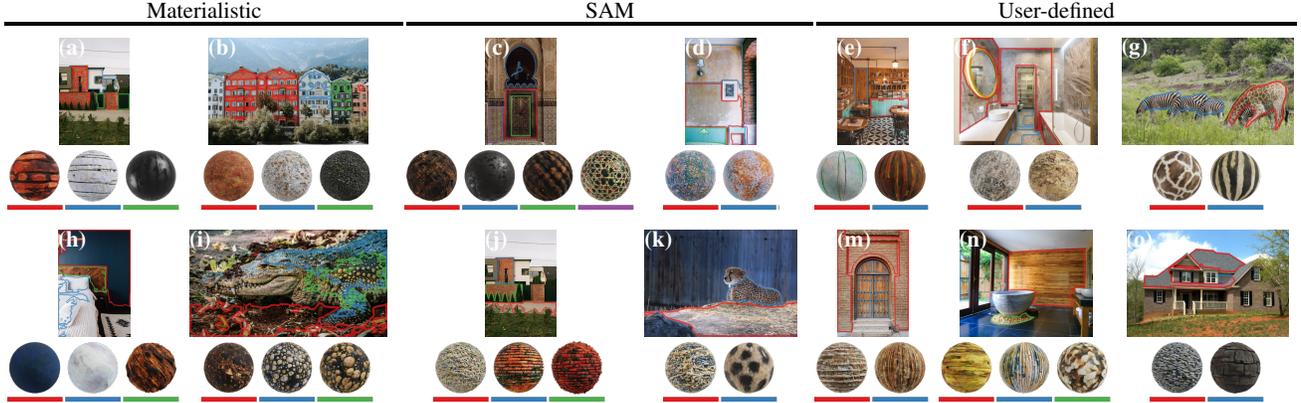

\begin{figure}[t]
	\centering
	\newcommand{\cred}[1]{\cellcolor{red!5}}
	\newcommand{\cgreen}[1]{\cellcolor{green!5}}
	\setlength{\tabcolsep}{0pt}
	\renewcommand{\arraystretch}{0}
	
	\resizebox{\columnwidth}{!}{
		\begin{tabular}{cc @{\hskip .4em} cc}

            \begin{tikzpicture}
                \node[anchor=south west, inner sep=0] (image) at (0,0) {\includegraphics[width=163px, trim=10px 15px 30px 0, clip, cframe=black .2px]{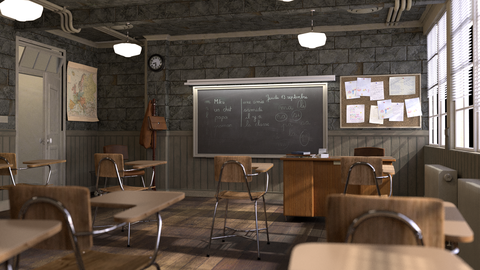}};
                \node[anchor=north west, inner sep=0] (overlay) at (image.north west) {\includegraphics[width=.2\columnwidth, trim=10px 6px 30px 0, clip]{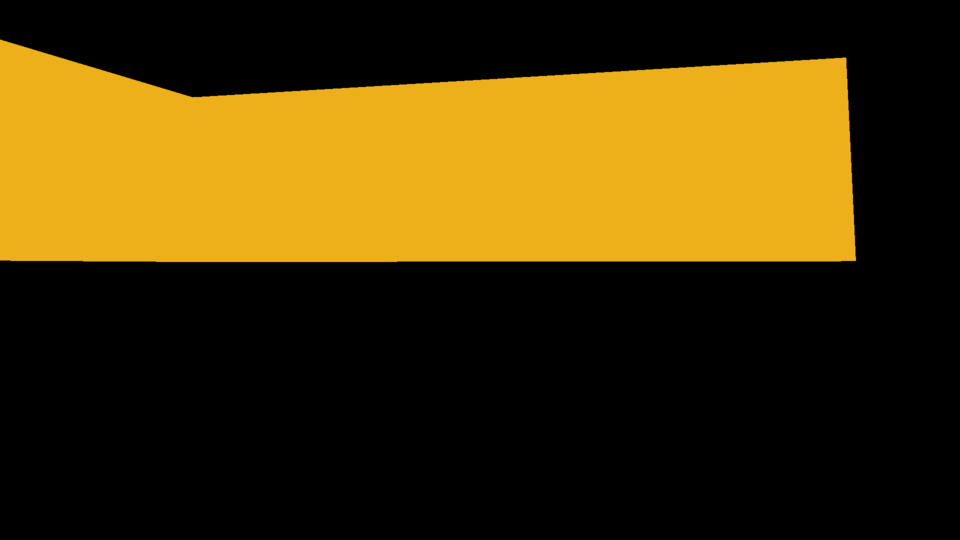}};
			\end{tikzpicture}&%
			\includegraphics[height=100px,cframe=black .2px]{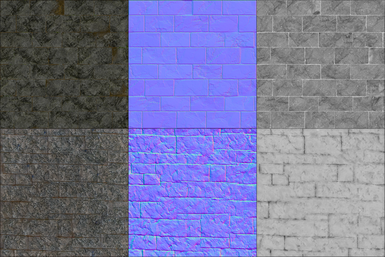}&%

            \begin{tikzpicture}
                \node[anchor=south west, inner sep=0] (image) at (0,0) {\includegraphics[width=163px, trim=10px 15px 30px 0, clip, cframe=black .2px]{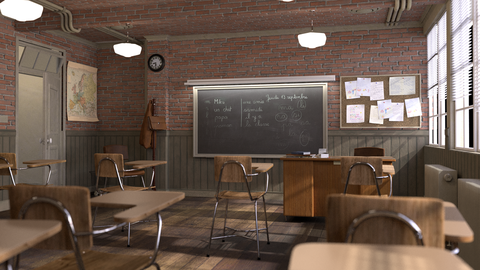}};
                \node[anchor=north west, inner sep=0] (overlay) at (image.north west) {\includegraphics[width=.2\columnwidth, trim=10px 6px 30px 0, clip]{images/end2end/mask_classroom_wall.png}};
			\end{tikzpicture}&%
			\includegraphics[height=100px,cframe=black .2px]{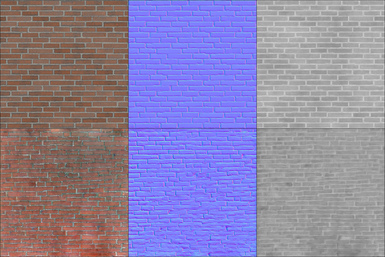}\\[2ex]

            \begin{tikzpicture}
                \node[anchor=south west, inner sep=0] (image) at (0,0) {\includegraphics[width=163px, cframe=black .2px]{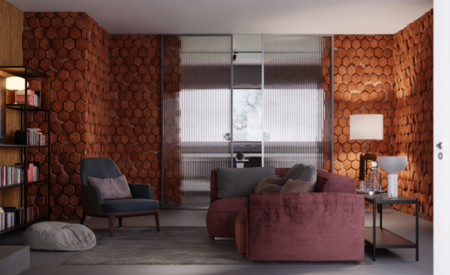}};
                \node[anchor=north west, inner sep=0] (overlay) at (image.north west) {\includegraphics[width=.2\columnwidth]{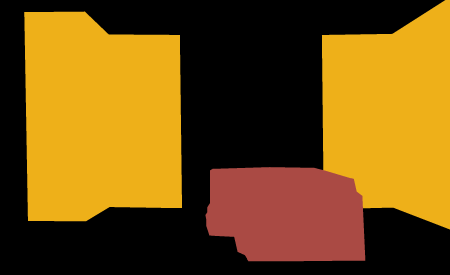}};
			\end{tikzpicture}&%
			\includegraphics[height=100px,cframe=black .2px]{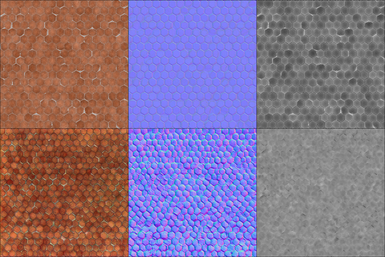}&%

            \begin{tikzpicture}
                \node[anchor=south west, inner sep=0] (image) at (0,0) {\includegraphics[width=163px, cframe=black .2px]{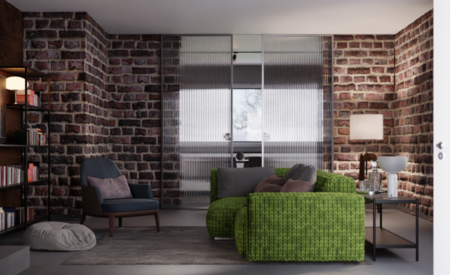}};
                \node[anchor=north west, inner sep=0] (overlay) at (image.north west) {\includegraphics[width=.2\columnwidth]{images/end2end/mask_living_wallsofa.png}};
			\end{tikzpicture}&%
			\includegraphics[height=100px,cframe=black .2px]{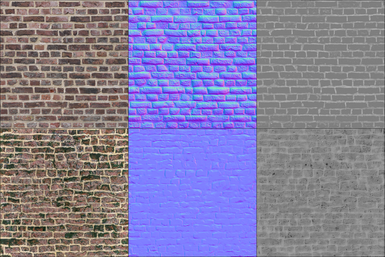}\\[1ex]

            \begin{tikzpicture}
                \node[anchor=south west, inner sep=0] (image) at (0,0) {\includegraphics[width=163px, cframe=black .2px]{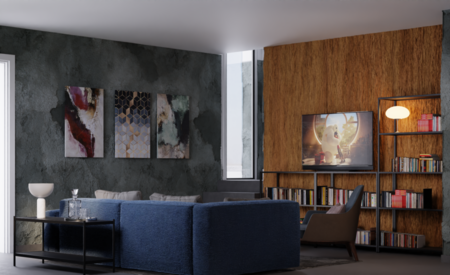}};
                \node[anchor=north west, inner sep=0] (overlay) at (image.north west) {\includegraphics[width=.2\columnwidth]{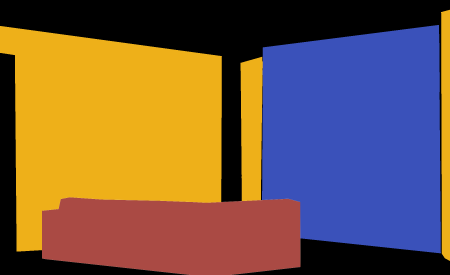}};
			\end{tikzpicture}&%
			\includegraphics[height=100px,cframe=black .2px]{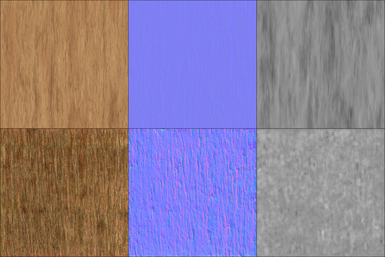}&%

			\multicolumn{2}{c}{
				{
					\raisebox{5em}{
						\setlength{\tabcolsep}{7pt}
						\renewcommand{\arraystretch}{0.85}
						\resizebox{0.6\textwidth}{!}{
							\begin{tabular}{c|ccc}
								\toprule
								&\multicolumn{3}{c}{LPIPS $\downarrow$}\\
								$P$ & \textbf{A} & \textbf{N} & \textbf{R}\\
								\toprule
								SD rand. & 0.8163	& 0.6895 & 0.7441 \\
								ours (user) & \textbf{0.6839} & \textbf{0.6254} & \textbf{0.5951} \\  
								\midrule
								\cgreen{}ACG rand. & \cgreen{}0.6547 & \cgreen{}0.4508 & \cgreen{}0.5600 \\
								\bottomrule
							\end{tabular}
						}
					}
				}
			}
		\end{tabular}%
	}

	\caption{\textbf{End to end evaluation.} We edit 3D scenes with \acg materials rendering different scenes and viewpoints. This allows us to compare \method{} extractions w.r.t. \acg ground truth, and quantify our end-to-end performance (extraction $+$ decomposition). Extracted regions are shown in top-left insets. %
	}%
	\label{fig:acg2render} %
\end{figure}

\subsection{Material extraction}
\label{sec:exp-material}
Given the lack of datasets combining real scenes with region-wise materials annotations, we highlight the complexity of evaluating all components of our method together. %

\condenseparagraph{Resemblance to \svbrdf{}dataset.} We design experiments using \os (\OS) and \acg, allowing us to {understand if \method preserves the expected characteristics of a material region.}
Considering that datasets class ontologies differ, we map \OS and \acg classes to a common set of 14 materials $C_m$, detailed in the supplementary, in which \OS classes are grouped following~\cite{xiao2018unified}. 

In a first experiment, we use \OS ground truth, \ie{} user-annotated material segmentation masks, as~$\mathcal{R}$, and automatically extract the associated materials $M_\text{SD}$ with~\method. Given that in the \OS ground truth the material class $c$ of $\mathcal{R}$ is known (but not the intrinsics), we compare the extracted $M_\text{SD}$ with those of materials of the same class in~\acg. 
We define an \textit{upper bound} by evaluating the same extracted materials but on \textit{all other materials} in~\acg. Intuitively, if results are better than the upper bound, we correctly mapped the appearance of a particular material class to visual features specific to that class. 
In other words, a material labeled as `brick' in a natural image should lead to extracted maps more similar to `brick' samples in \acg, than to other classes such as `wood'. In practice, the evaluation is conducted by sampling 100 $\{M_\text{SD}, \tilde{M}\}$ pairs and evaluating \lpips~\cite{zhang2018unreasonable} between them (lower is better), for each of the 14 material classes. 
Considering class $c \in C_m$, \textit{Ours} will evaluate $\{M_\text{SD}^c, \tilde{M}^c\}$ pairs $\forall c \in C_m$, while the \textit{upper bound} will have $\{M_\text{SD}^c, \tilde{M}^{\bar{c}}\}$ where $\bar{c}$ is a random ${\bar{c}} \neq c$. The reported LPIPS values are averaged over all classes. 
From Tab.~\ref{tab:quant-material-extraction}~(left), \mbox{`Ours-\OS Masks'} improves performance over the upper bound, proving the effectiveness of our method. 
Additionally, we propose a \textit{lower bound}, following our pipeline but using $\{M_\text{ACG}^c, \tilde{M}^c\}$ where $M_\text{ACG}$ are estimated ACG maps from single-view samples of class $c$. We also evaluate the impact on material extraction of our automated pipeline using either SAM~\cite{kirillov2023segment} or Materialistic~\cite{sharma2023materialistic} as region segmenters (Sec.~\ref{sec:automation}). We highlight how in all setups we achieve comparable performance, always improving over the lower bound. %

{In a second experiment, we propose an evaluation with CLIP~\cite{radford2021learning}. We combine all generated textures of $c \in C_m$ with \promptsMaterial (\cf Sec.~\ref{sec:exp-ablation}) and evaluate the CLIP ViT-B/32 zero-shot classification performance of all rendered $M_\text{SD}$ with random illumination. We do the same for all ground truth $\tilde{M}$ in \acg. In Tab.~\ref{tab:quant-material-extraction} (right) we report average comparable accuracies, suggesting that we successfully render similar materials to existing PBR datasets, $\forall c \in C_m$.}

\condenseparagraph{Qualitative evaluation.} In~\cref{fig:palettes} we visualize web-scraped images and the materials extracted by \method{} using regions from SAM~\cite{kirillov2023segment}, Materialistic~\cite{sharma2023materialistic} or a User input. 
Each material is rendered on a 3D sphere in Blender with a color below matching its region color. %
Given the task complexity, we emphasize the quality of the extraction for a wide variety of complex materials: bricks, tiles, skins, fur, \etc.
In particular, we highlight the quality of bricks in \mathi{palred}{\textbf{(a)}}\footnote{Here, \mathi{palred}{\textbf{(a)}} refers to material of the \mathi{palred}{red} region from image \textbf{(a)} of~\cref{fig:palettes}.} as well as in \mathi{palblue}{\textbf{(j)}} and \mathi{palgreen}{\textbf{(j)}} with a different segmenter, and more astonishingly in the small roof region of \mathi{palblue}{\textbf{(o)}}. In natural images, materials also match appearances such as crocodile skin~\mathi{palgreen}{\textbf{(i)}}, or fur of jaguar \mathi{palblue}{\textbf{(k)}}, giraffe \mathi{palred}{\textbf{(g)}} and zebra \mathi{palblue}{\textbf{(g)}}. Other noticeable results are the complex mosaic pattern \mathi{palpurple}{\textbf{(c)}} or damaged wall \mathi{palred}{\textbf{(d)}} and \mathi{palblue}{\textbf{(d)}}.

\condenseparagraph{End-to-end re-rendering.} We also design a challenging end-to-end evaluation leveraging realistic 3D scenes. Through automatic edition we replace some 3D objects material with a PBR material of \acg, thus rendering a total of 174 images (2 scenes, 4 views) comprising 10 materials for each of the 16 dominant classes of~\acg.
As the latter comes with SVBRDF ground truths, we compare them with our \method{} extractions on the rendered images with \textit{ad-hoc} user-input regions. 
Visuals in~\cref{fig:acg2render} show side-by-side renderings, each with \acg ground truth (top) and our extractions (bottom). Our materials capture the main characteristics though we observe some over/under saturation, highlighting the task complexity. Moreover, table inset in~\cref{fig:acg2render} compares \lpips ($\downarrow$) vs ground truth \acg for either a `SD random' generation \textit{prompted with the true class}, or a random \acg material of the true class. Notably, our materials are much closer to \acg than SD generation, demonstrating the benefit of our pipeline over text-to-image.

\begin{figure}[t]
	\definecolor{mediumtealblue}{rgb}{0.0, 0.33, 0.71}
	\definecolor{mediumcandyapplered}{rgb}{0.89, 0.02, 0.17}
	\definecolor{mediumseagreen}{rgb}{0.24, 0.7, 0.44}
	\definecolor{mordantred19}{rgb}{0.68, 0.05, 0.0}
	
	\definecolor{darkgreen}{RGB}{0,150,0}
	\definecolor{darkred}{RGB}{150,0,0}
	\definecolor{darkblue}{RGB}{0,0,150}
	\newcommand{\ored}[1]{\cellcolor{darkred!60}}
	\newcommand{\ogreen}[1]{\cellcolor{darkgreen!60}}
	\newcommand{\oblue}[1]{\cellcolor{darkblue!60}}
	
	\definecolor{color1}{RGB}{0,150,150}
	\definecolor{color2}{RGB}{150,0,150}
	\definecolor{color3}{RGB}{150,150,0}
	\definecolor{color4}{RGB}{100,100,100}
	\newcommand{\cone}[1]{\cellcolor{color1!60}}
	\newcommand{\ctwo}[1]{\cellcolor{color2!60}}
	\newcommand{\cthree}[1]{\cellcolor{color3!60}}
	\newcommand{\cfour}[1]{\cellcolor{color4!60}}
	
	\centering
	\renewcommand{\arraystretch}{0.4}
	\setlength{\tabcolsep}{0.004\linewidth}
	\resizebox{.98\columnwidth}{!}{
		\begin{tabular}{c ccccccc}

			\multicolumn{1}{c}{\textbf{}} & region & \multicolumn{2}{c}{$c_{in}=128$} & \multicolumn{2}{c}{$c_{in}=256$} & \multicolumn{2}{c}{$c_{in}=512$} \\
			\cmidrule[1pt](lr){3-4}\cmidrule[1pt](lr){5-6}\cmidrule[1pt](lr){7-8}
			
			\multirow{1}{*}[3em]{\rotatebox{90}{$c_x=64$}}
			& \includegraphics[width=5em, height=5em]{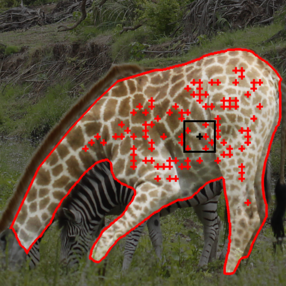} 
			& \includegraphics[width=5em, height=5em]{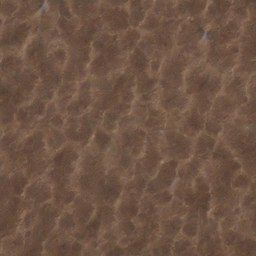}
			& \includegraphics[width=5em, height=5em]{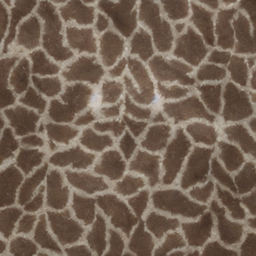}
			& \includegraphics[width=5em, height=5em]{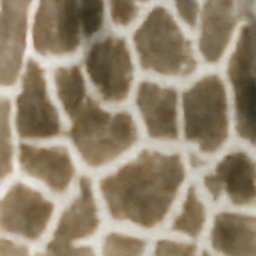} 
			& \includegraphics[width=5em, height=5em]{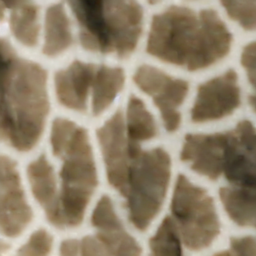} 
			& \includegraphics[width=5em, height=5em]{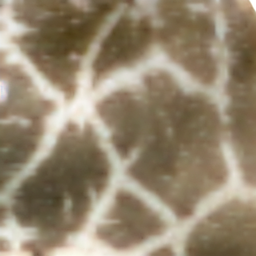}
			& \includegraphics[width=5em, height=5em]{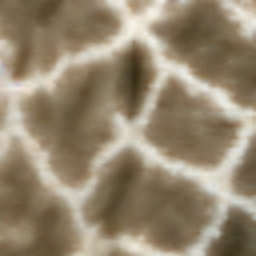} \\
			
			\multirow{1}{*}[3.2em]{\rotatebox{90}{$c_x=128$}}
			& \includegraphics[width=5em, height=5em]{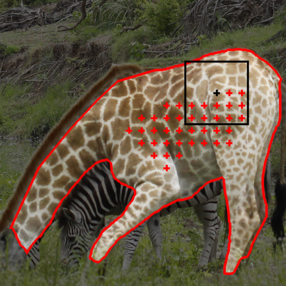} 
			& \includegraphics[width=5em, height=5em]{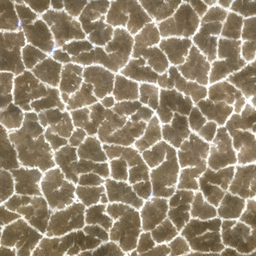} 
			& \includegraphics[width=5em, height=5em]{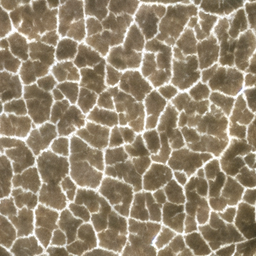} 
			& \includegraphics[width=5em, height=5em, cframe=c2 1pt]{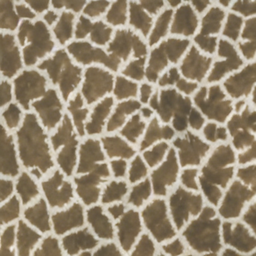} 
			& \includegraphics[width=5em, height=5em, cframe=c2 1pt]{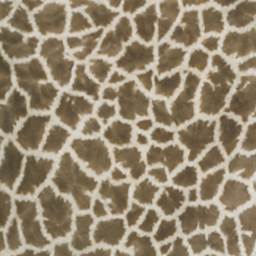} 
			& \includegraphics[width=5em, height=5em]{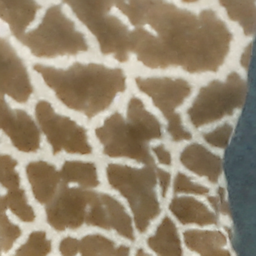}
			& \includegraphics[width=5em, height=5em]{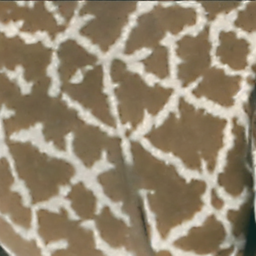} \\

			\bottomrule
			\vspace{0.5px}
		\end{tabular}
	}		
	
	\resizebox{.98\columnwidth}{!}{
		\begin{tabular}{c cccccccc}
			
			& \multicolumn{2}{c}{\footnotesize{\textit{``a photo of $\textbf{S}^{*}$ texture''}}} 
			& \multicolumn{2}{c}{\footnotesize{\textit{``$\textbf{S}^{*}$''}}} 
			& \multicolumn{2}{c}{\footnotesize{\textit{``a photo of a $\textbf{S}^{*}$''}}} 
			& \multicolumn{2}{c}{\footnotesize{\textit{``an object with $\textbf{S}^{*}$ texture''}}} \\
			
			\cmidrule[1pt](lr){2-3}\cmidrule[1pt](lr){4-5}\cmidrule[1pt](lr){6-7}\cmidrule[1pt](lr){8-9}
			
			\multirow{1}{*}[2.5em]{\conesquare}
			
			& \includegraphics[width=5em, height=5em]{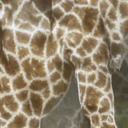}
			& \includegraphics[width=5em, height=5em]{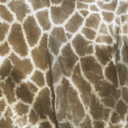}
			
			& \includegraphics[width=5em, height=5em]{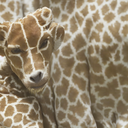}
			& \includegraphics[width=5em, height=5em]{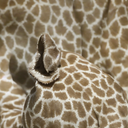}
			
			& \includegraphics[width=5em, height=5em]{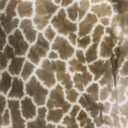}
			& \includegraphics[width=5em, height=5em]{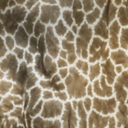}
			& \includegraphics[width=5em, height=5em]{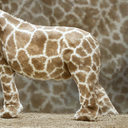}
			& \includegraphics[width=5em, height=5em]{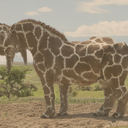} \\
			
			\multirow{1}{*}[2.5em]{\cfoursquare}
			
			& \includegraphics[width=5em, height=5em]{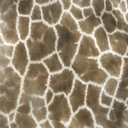}
			& \includegraphics[width=5em, height=5em]{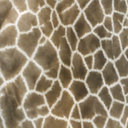}
			
			& \includegraphics[width=5em, height=5em]{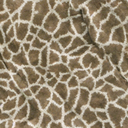}
			& \includegraphics[width=5em, height=5em]{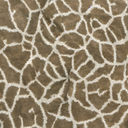}
			
			& \includegraphics[width=5em, height=5em]{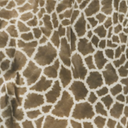}
			& \includegraphics[width=5em, height=5em]{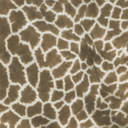}
			& \includegraphics[width=5em, height=5em, cframe=c2 1pt]{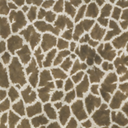}
			& \includegraphics[width=5em, height=5em, cframe=c2 1pt]{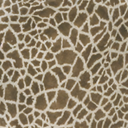} \\
			
			\multirow{1}{*}[2.5em]{\cfivesquare}
			
			& \includegraphics[width=5em, height=5em]{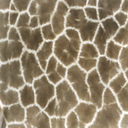}
			& \includegraphics[width=5em, height=5em]{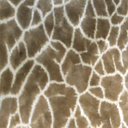}
			
			& \includegraphics[width=5em, height=5em]{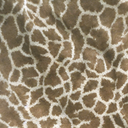}
			& \includegraphics[width=5em, height=5em]{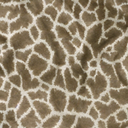}
			
			& \includegraphics[width=5em, height=5em]{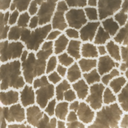}
			& \includegraphics[width=5em, height=5em]{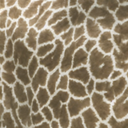}
			& \includegraphics[width=5em, height=5em, cframe=c2 1pt]{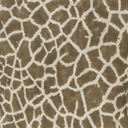}
			& \includegraphics[width=5em, height=5em, cframe=c2 1pt]{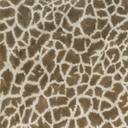} \\
			
		\end{tabular}
	}\vspace{-0.2em}
	
	\caption{\textbf{Inversion ablation.} We show how crop and upsampling sizes affect SD generation results (top). Effects of \promptTrain for learning $\textbf{S}^{*}$ are also given (bottom). For all four inversions, we provide generations using the templates \conesquare, \cfoursquare, and \cfivesquare~(\cf \cref{fig:prompt-ablation}). We highlight in {\color{c2} green} the parameters used.
    }
	\label{fig:prompt-learning}
\end{figure}

\begin{figure}[t]
	\centering
	\setlength{\tabcolsep}{0.005\linewidth}
	\resizebox{\columnwidth}{!}{
		\begin{tabular}{cl|cccH|cccH}
			&&\multicolumn{4}{c|}{FID $\downarrow$}&\multicolumn{4}{c}{KID $\downarrow$}\\
			& Prompt templates   & \textbf{A} & \textbf{N} & \textbf{R} & I   & \textbf{A} & \textbf{N} & \textbf{R} & I  \\
			\midrule
			\conesquare   & \small{\textit{``a photo of a $\textbf{S}^{c}$''}}                                  & 1.89           & 1.74           & 1.83           & 1.83              & 2.60           & 6.52           & 3.75           & 2.89              \\
			\ctwosquare   & \small{\textit{``a $\textbf{S}^{c}$ material''}}                                    & 1.83           & 1.60           & 1.67           & 1.73              & 2.38           & 5.42           & \textbf{2.74}           & 2.66              \\
			\cthreesquare & \small{\textit{``a $\textbf{S}^{c}$ texture''}}                                     & 1.72           & 1.54           & 1.63           & 1.68              & 1.82           & 5.27           & 2.86           & 2.53              \\
			\cfoursquare  & \small{\textit{``realistic $\textbf{S}^{c}$ texture in top view''}}                 & \textbf{1.55}  & \textbf{1.27}  & 1.59           & \textbf{1.50}     & \textbf{1.54}  & \textbf{3.34}  & 3.32  & \textbf{1.93}     \\
			\cfivesquare  & \small{\textit{``high resolution realistic $\textbf{S}^{c}$ texture in top view''}} & \textbf{1.55}  & 1.30           & \textbf{1.55}  & \normalsize{1.53} & 1.71           & 3.48           & 3.34           & \normalsize{2.08} \\
			\bottomrule
			\vspace{0.2px}
		\end{tabular}%
	}
	
	\centering
	\setlength{\tabcolsep}{0.004\linewidth}
	\resizebox{\columnwidth}{!}{
		
		\begin{tabular}{c ccccc|c}
			
			$\textbf{S}^{c}$ & \conesquare & \ctwosquare & \cthreesquare & \cfoursquare & \cfivesquare & \acg \\ \midrule
			\multirow{1}{*}[2.5em]{\rotatebox{90}{Paper}}
			
			& \includegraphics[width=5em, height=5em]{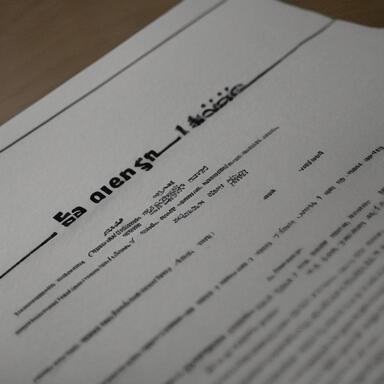} 
			& \includegraphics[width=5em, height=5em]{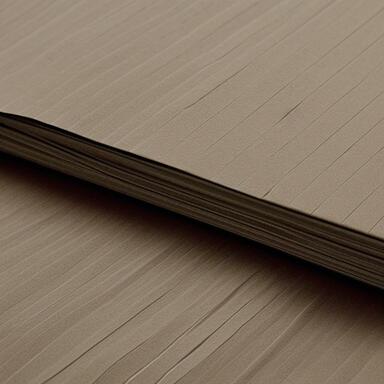} 
			& \includegraphics[width=5em, height=5em]{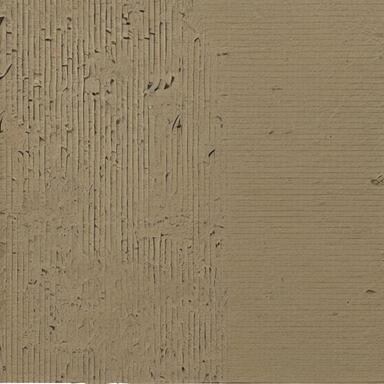} 
			& \includegraphics[width=5em, height=5em]{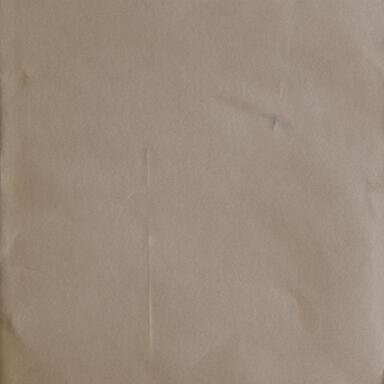} 
			& \includegraphics[width=5em, height=5em]{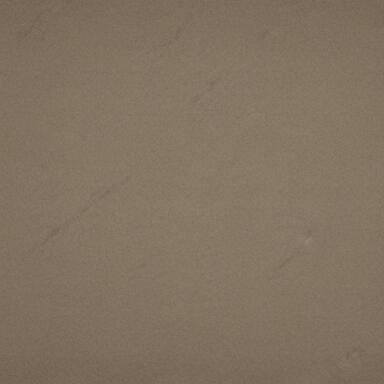} 
			& \includegraphics[width=5em, height=5em]{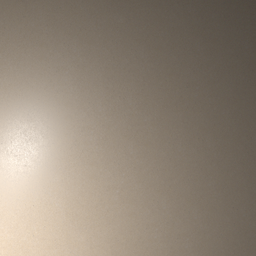}\\ 
			{\rotatebox{90}{Pav. Stones}}
			& \includegraphics[width=5em, height=5em]{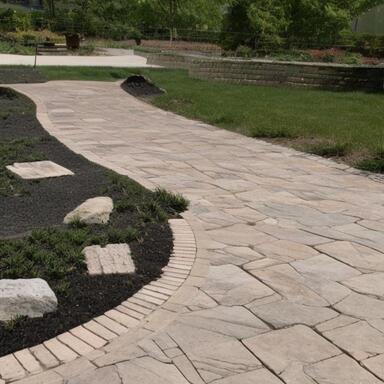} 
			& \includegraphics[width=5em, height=5em]{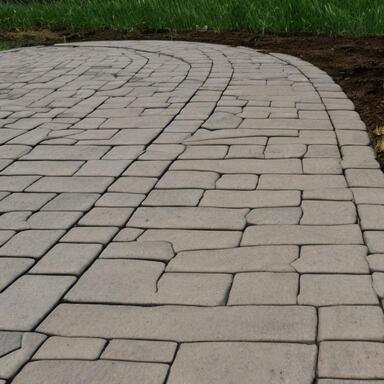} 
			& \includegraphics[width=5em, height=5em]{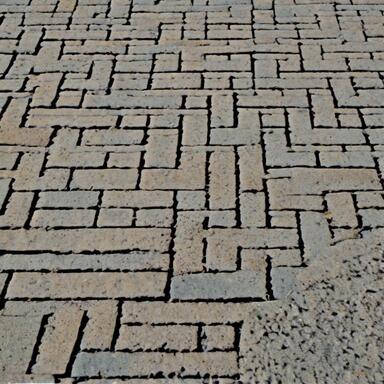} 
			& \includegraphics[width=5em, height=5em]{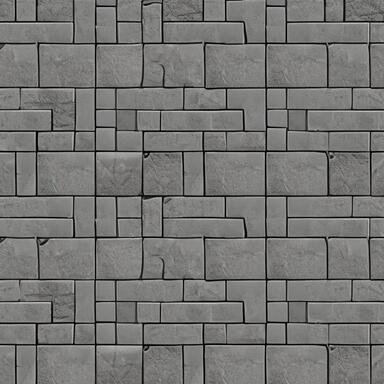} 
			& \includegraphics[width=5em, height=5em]{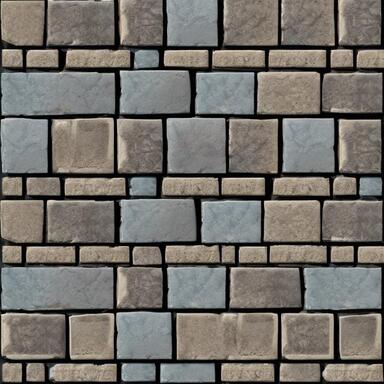} 
			& \includegraphics[width=5em, height=5em]{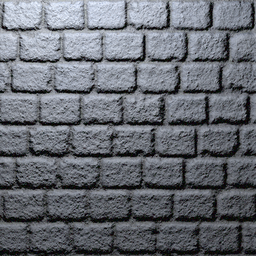}\\ 
		\end{tabular}
	}\vspace{-0.1em}
	
	\caption{\textbf{Prompt generation.} We compare prompt templates when generating using class names as $\textbf{S}^{c}$. With a more detailed prompt and the word ``texture'', we are able to generate images similar to the ones in~\acg. We report FID $\cdot 10^{-2}$ and KID $\cdot 10^{2}$.}%
	\label{fig:prompt-ablation}
\end{figure}

\begin{figure}[t]
	\centering
	\tiny
	\setlength{\tabcolsep}{0pt}
	\renewcommand{\arraystretch}{0}
	\resizebox{\columnwidth}{!}{
		\begin{tabular}{cccc cccc cccc}
			\multicolumn{4}{c}{``classroom''} & \multicolumn{4}{c}{``flat-front''} & \multicolumn{4}{c}{``flat-back''} \\
			\cmidrule[.5pt](lr){1-4} \cmidrule[.5pt](lr){5-8} \cmidrule[.5pt](lr){9-12}

			\multicolumn{4}{c}{
				\begin{tikzpicture}
					\node[anchor=south west, inner sep=0] (image) at (0,0) {\includegraphics[width=.333\columnwidth, trim=10px 10px 30px 0, clip, cframe=black .2px]{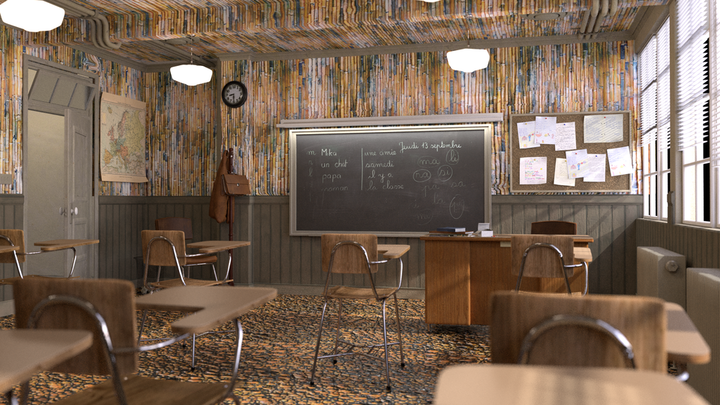}};
					\node[anchor=north west, inner sep=0] (overlay) at (image.north west) {\includegraphics[width=.1\columnwidth, trim=10px 10px 30px 0, clip]{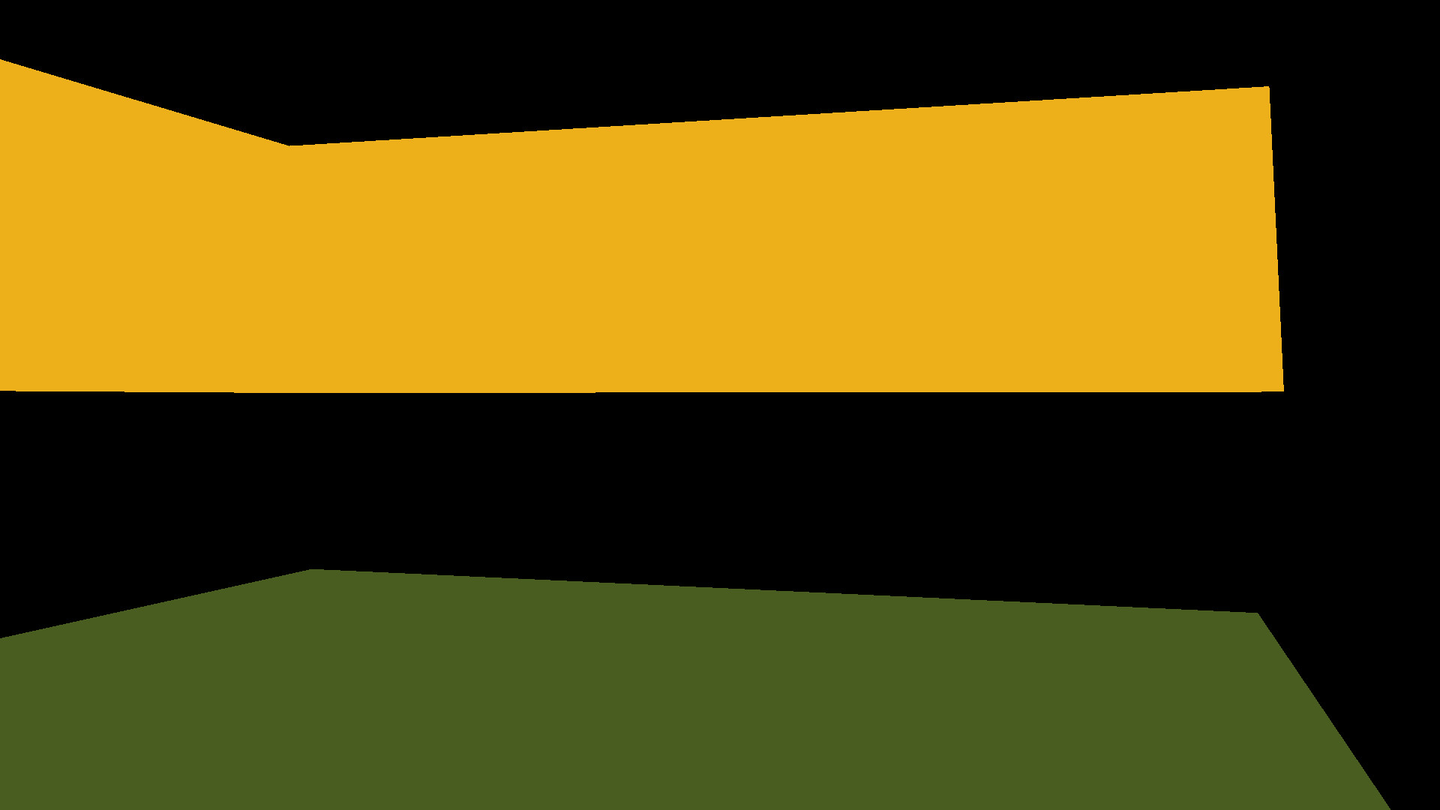}};
			\end{tikzpicture}}&%

			\multicolumn{4}{c}{
				\begin{tikzpicture}
					\node[anchor=south west, inner sep=0] (image) at (0,0) {\includegraphics[width=.333\columnwidth, cframe=black .2px]{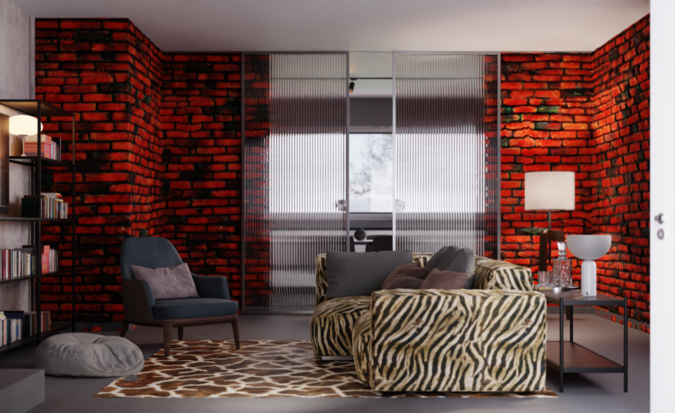}};
					\node[anchor=north west, inner sep=0] (overlay) at (image.north west) {\includegraphics[width=.1\columnwidth]{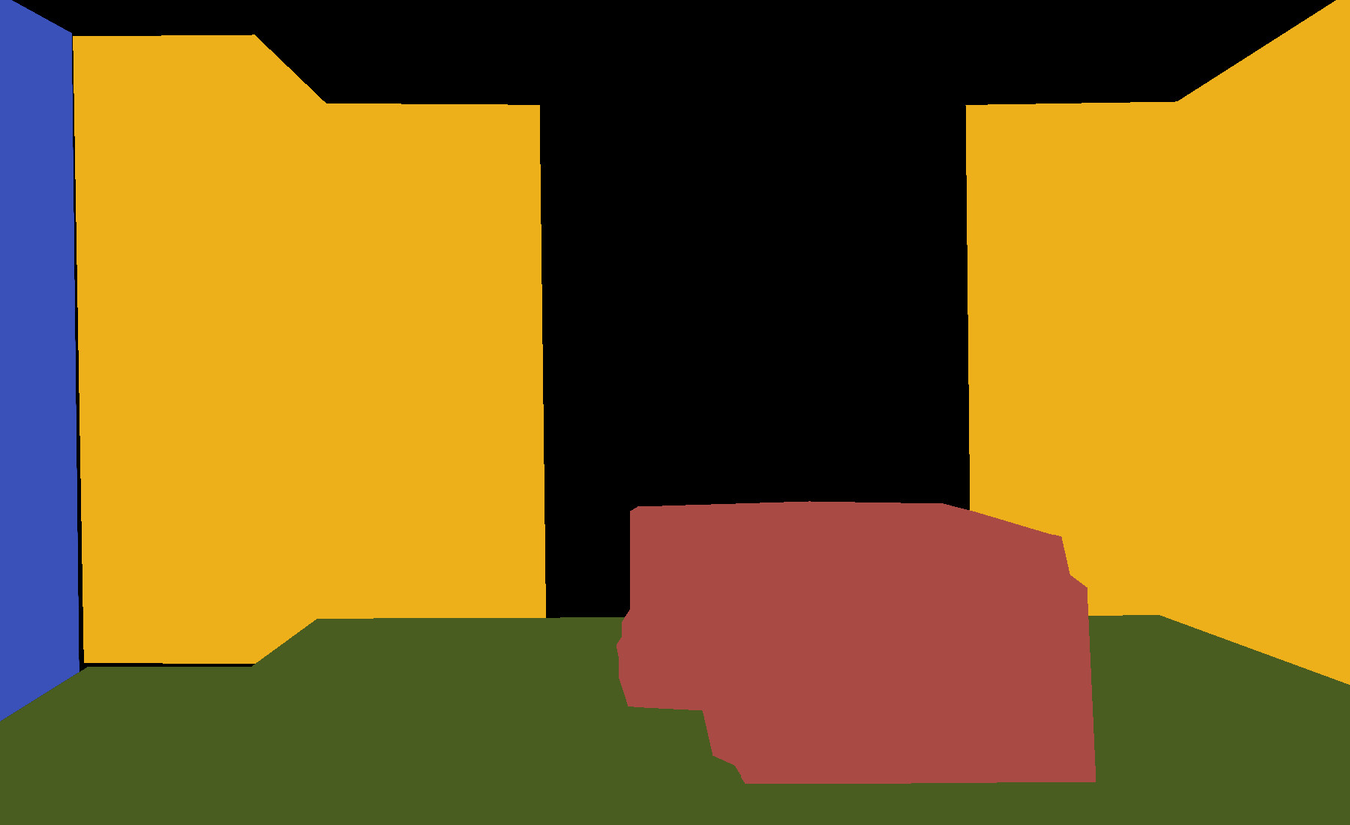}};
			\end{tikzpicture}}&%

			\multicolumn{4}{c}{
				\begin{tikzpicture}
					\node[anchor=south west, inner sep=0] (image) at (0,0) {\includegraphics[width=.333\columnwidth, cframe=black .2px]{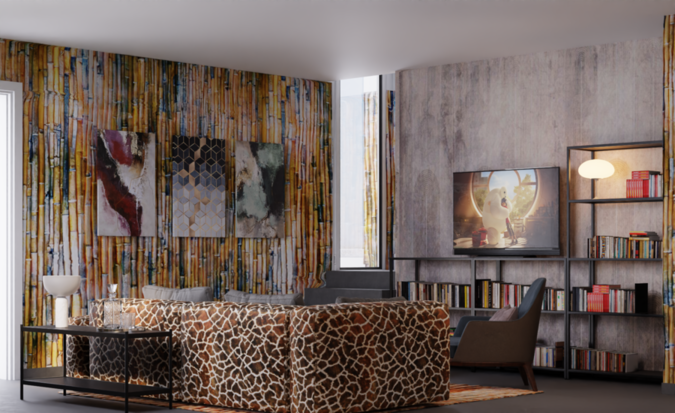}};
					\node[anchor=north west, inner sep=0] (overlay) at (image.north west) {\includegraphics[width=.1\columnwidth]{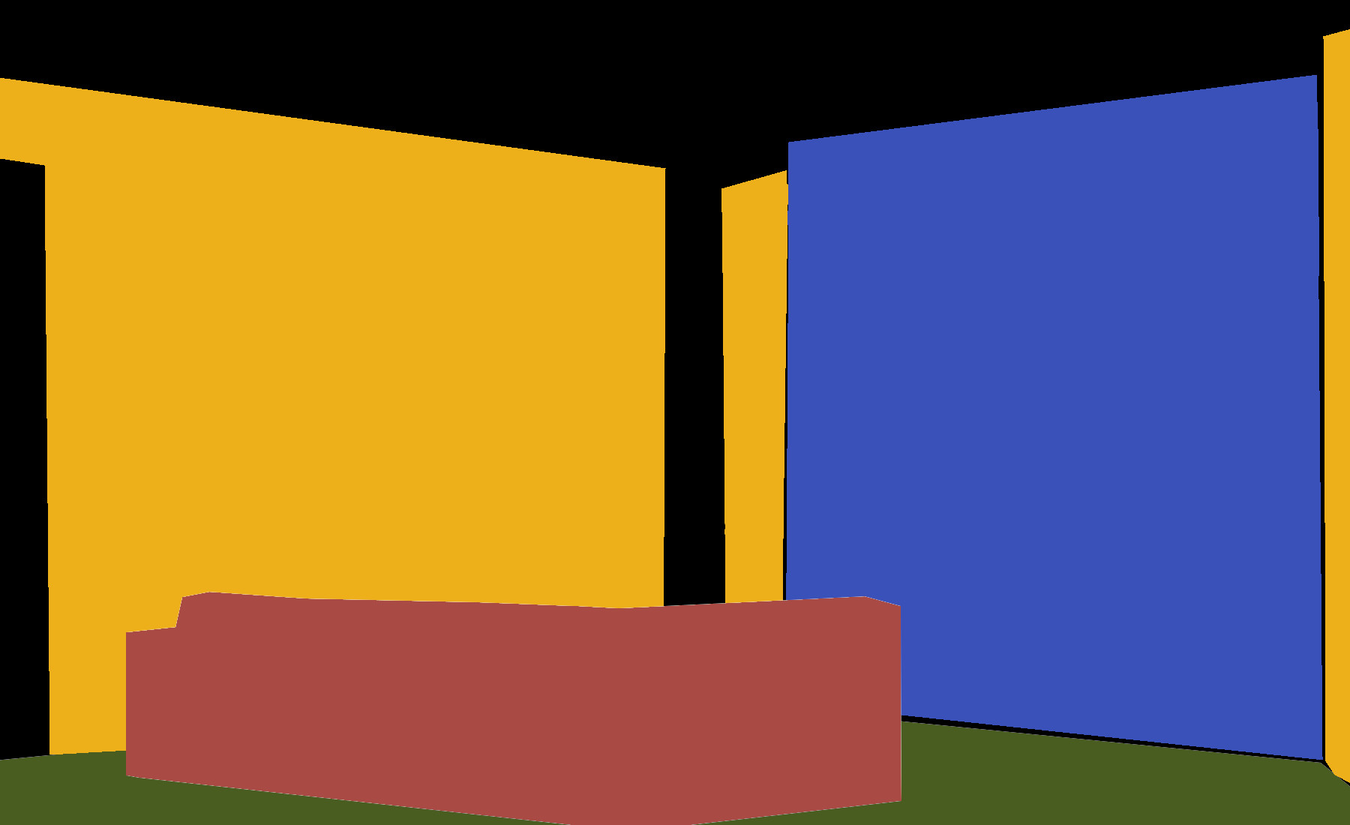}};
			\end{tikzpicture}}\\
			
			\multicolumn{2}{c}{\includegraphics[width=.1666\columnwidth, trim=10px 6px 30px 0, clip, cframe=black .1px]{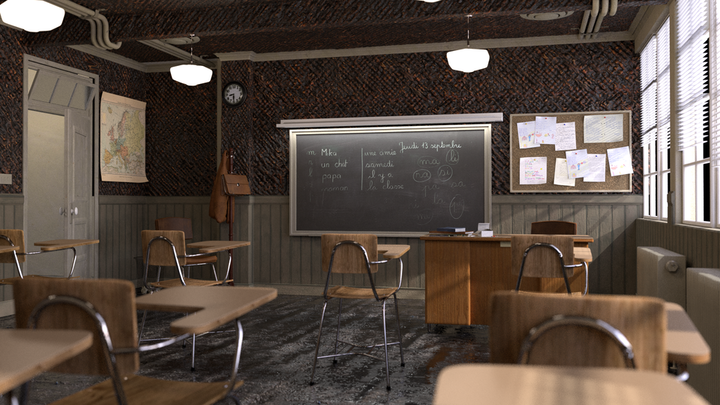}} &%
			\includegraphics[width=.1666\columnwidth, trim=10px 6px 30px 0, clip, cframe=black .1px]{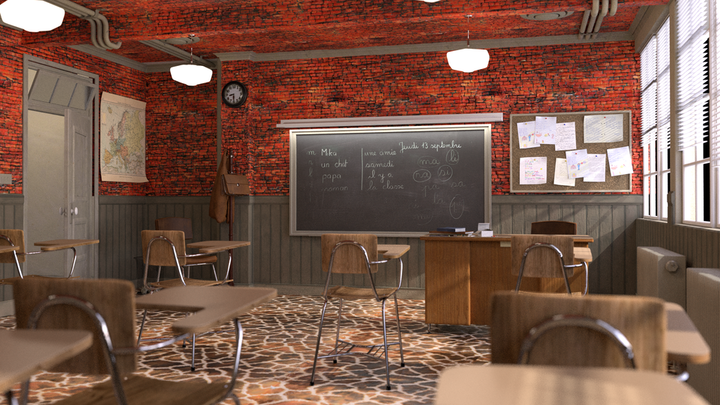} &
			\multicolumn{2}{c}{\includegraphics[width=.1666\columnwidth, cframe=black .1px]{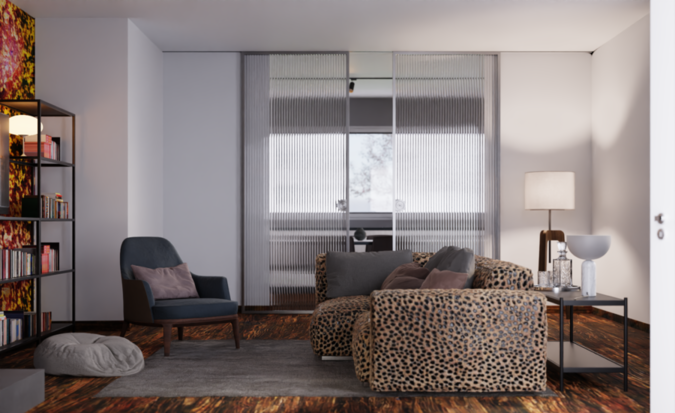}} &%
			\multicolumn{2}{c}{\includegraphics[width=.1666\columnwidth, cframe=black .1px]{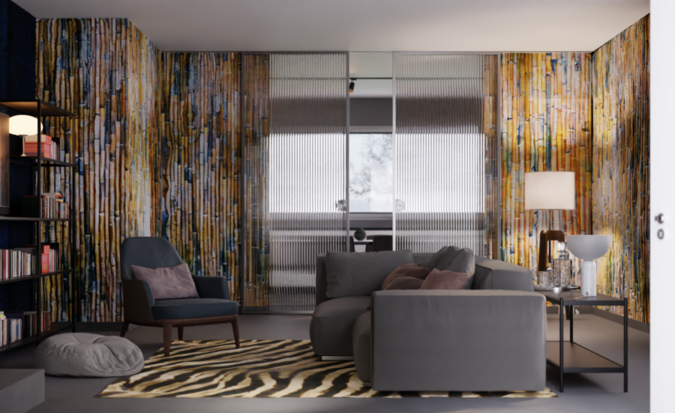}} &%
			\multicolumn{2}{c}{\includegraphics[width=.1666\columnwidth, cframe=black .1px]{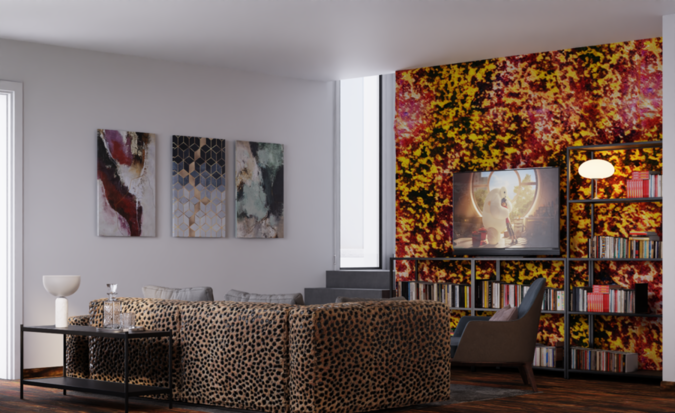}} &%
			\multicolumn{2}{c}{\includegraphics[width=.1666\columnwidth, cframe=black .1px]{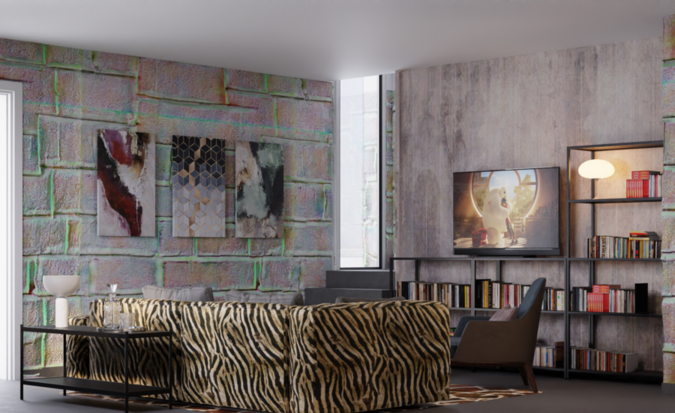}} \\
			
			\multicolumn{2}{c}{\includegraphics[width=.1666\columnwidth, trim=10px 6px 30px 0, clip, cframe=black .1px]{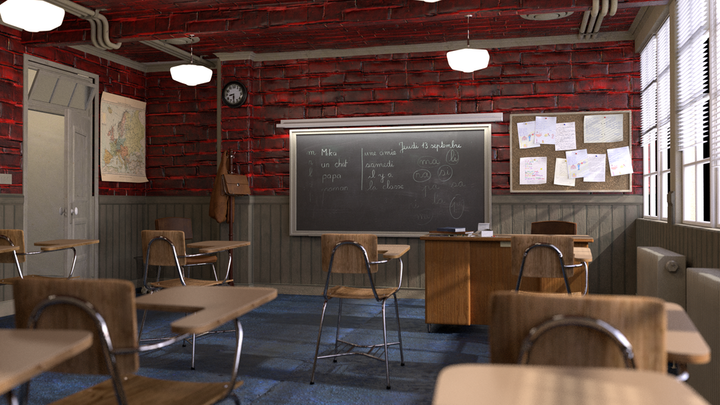}} &%
			\multicolumn{2}{c}{\includegraphics[width=.1666\columnwidth, trim=10px 6px 30px 0, clip, cframe=black .1px]{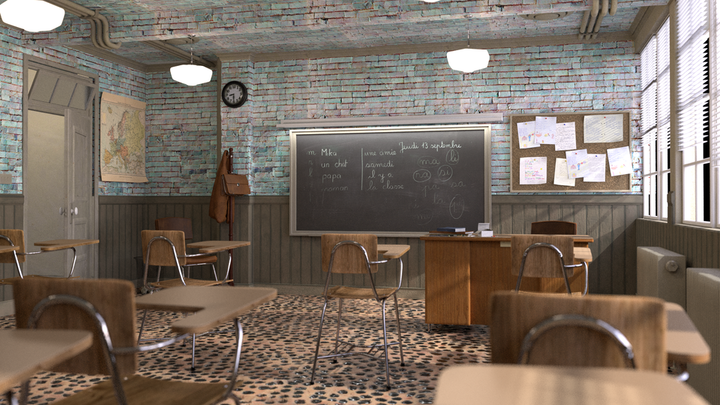}} &%
			\multicolumn{2}{c}{\includegraphics[width=.1666\columnwidth, cframe=black .1px]{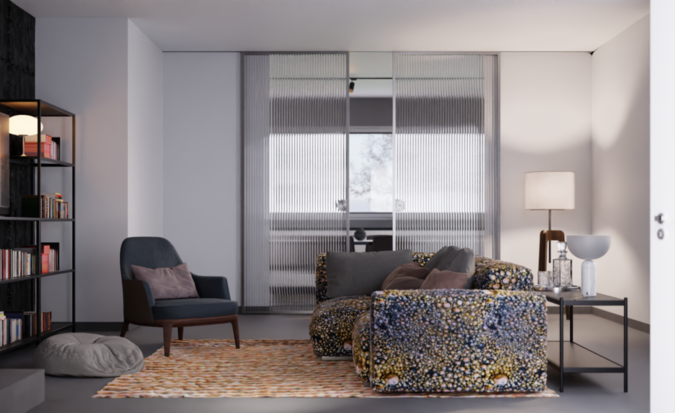}} &%
			\multicolumn{2}{c}{\includegraphics[width=.1666\columnwidth, cframe=black .1px]{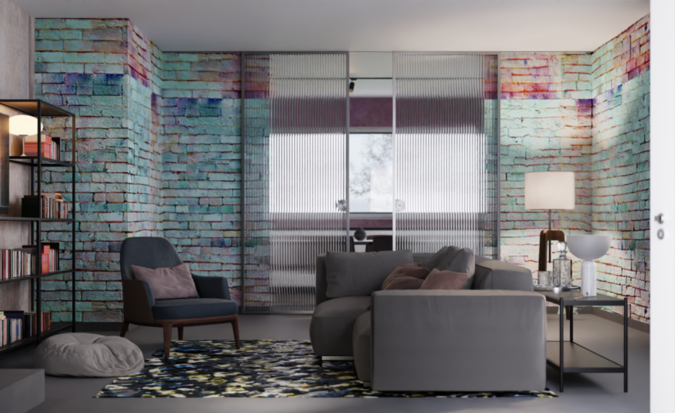}} &%
			\multicolumn{2}{c}{\includegraphics[width=.1666\columnwidth, cframe=black .1px]{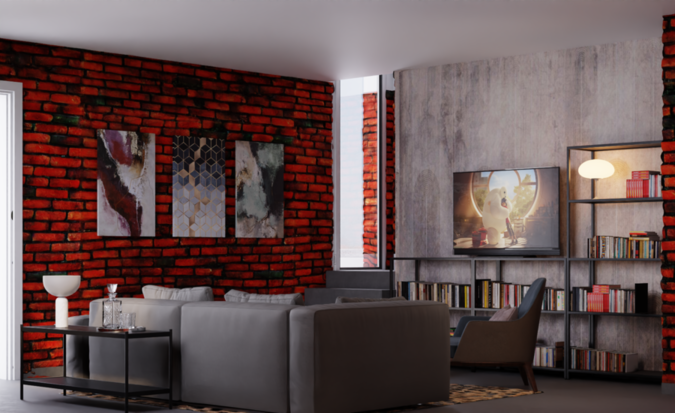}} &%
			\multicolumn{2}{c}{\includegraphics[width=.1666\columnwidth, cframe=black .1px]{images/ours_3d/bathroom_user_3d_flat-archiviz_wall-c4_sofa-c1_carpet-c1_1800x1100Cam_Sofa_Back.png}} \\
			
		\end{tabular}%
	}
	\caption{\textbf{Scene editing.} We edit objects (top-left insets) of 3D scenes using materials \textit{extracted from real-world images}.
	}%
	\label{fig:matex2render} %
    \vspace{-2em}
\end{figure}

\subsection{Ablations}
\label{sec:exp-ablation}

\condenseparagraph{Inversion.} 
Considering we face objects of varying sizes, we motivate our need for scale adaptive crop extraction. 
In \cref{fig:prompt-learning} (top), we show the crop size $c_x$ and training input size $c_{in}$ (\ie, upsampling size).
Considering that crops have a much lower resolution than the pre-trained SD v1.5 inputs (512px), we make two choices: (i) extract crops with largest $c_x$ possible within~\region, and (ii) finetune SD at a lower resolution ($c_{in}=256$). This minimizes the input distortion while retaining good generation at 512px and beyond.

Additionally, we evaluate~\cref{fig:prompt-learning} (bottom) the choice of~\promptTrain when learning~\concept by showing generations using three prompts from~\cref{fig:prompt-ablation}. While choosing \promptTrain (row-wise) plays less than \promptsMaterial,\FPc{what does this mean guys?} we use \textit{``an object with $\textbf{S}^{*}$ texture''} when training. This motivates our choices for learning~\concept (\cref{sec:generative}).

\condenseparagraph{Prompt engineering.} We find that choosing the correct \promptsMaterial allows generating material images with the correct appearance. We ablate in~\cref{fig:prompt-ablation} (top) different prompts by sampling 10 images per~\acg{} class and processing them with our decomposition (\cref{sec:uda}). We then evaluate FID and KID against \acg annotations and rendered images. We find that the word ``texture'' improves synthesis over generic templates and removes additional context favoring top-view appearance. Furthermore, the text-to-image generation benefits from additional adjectives. Visual comparison of generated samples is in~\cref{fig:prompt-ablation} (bottom).

\subsection{3D Scene editing.}
\label{sec:exp-app}

We consider the extracted materials for scene editing applications. In~\cref{fig:matex2render}, we present renderings of 3D scenes, replacing materials of objects (highlighted in insets) with ones extracted in \textit{real-world} images with \method. Note the realism of our jaguar (middle) and giraffe (right) sofas, or the bamboo wall (left).

\vspace{-.5em}
\section{Discussion}
\label{sec:conclusion}
We introduced ~\method, a comprehensive approach designed to extract tileable, high-resolution PBR materials from single real-world images. Although capable of extracting accurate materials, our method faces some unexpected limitations. 
For example, while prior methods may struggle at regressing complex patterns we found it more challenging to capture simple uniform materials. In such cases, the concept collapses, leading to color artifacts, common in diffusion models. 
Another more predictable issue, involves illumination ambiguities -- particularly noticeable in shaded surfaces -- causing inconsistent colors. Lastly, \method is capable of making some geometric corrections, but cannot rectify slanted surfaces or account for strong distortion (perspective, lenses, depth of field). Addressing these shortcomings calls for further refinements. \method shows very promising results on a newly introduced challenging task. We hope our work sparks interesting research in the same direction.

\vspace{.1em}\noindent{}{\footnotesize \textbf{Acknowledgment.} This research project was funded by the French project SIGHT (ANR-20-CE23-0016). It was performed using HPC resources from GENCI–IDRIS (Grant 2023-AD011014389).}

{\small
	\bibliographystyle{ieeenat_fullname}
	\bibliography{biblio}
}

\end{document}